\documentclass{article}

\usepackage{microtype}
\usepackage{graphicx}
\usepackage{booktabs} 

\usepackage{hyperref}

\usepackage{url}
\usepackage{enumitem}
\usepackage{algpseudocode}
\usepackage{amsmath}  
\usepackage{amssymb}  
\usepackage{multirow}  
\usepackage{xcolor}
\usepackage{subcaption} 
\usepackage{float}
\usepackage{pifont}     
\usepackage{makecell}

\usepackage{listings}
\newcommand{\algname}[1]{BLASST}

\DeclareMathOperator{\rowmax}{rowmax}
\DeclareMathOperator{\rowsum}{rowsum}

\newcommand{\cmark}{\ding{51}}
\newcommand{\xmark}{\ding{55}}

\usepackage{siunitx}
\usepackage[accepted]{mlsys2026}

\AddToShipoutPictureBG*{
\AtPageUpperLeft{
  \hspace*{\paperwidth}
  \raisebox{-68pt}{
    \llap{
      \href{https://www.acm.org/publications/policies/artifact-review-and-badging-current}{
        \includegraphics[height=65pt]{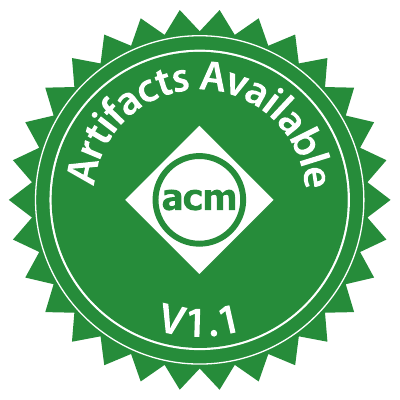}}
      \hspace{1pt}
      \href{https://www.acm.org/publications/policies/artifact-review-and-badging-current}{
        \includegraphics[height=65pt]{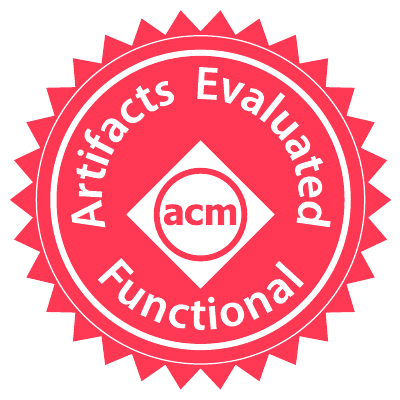}}
      \hspace{1pt}
      \href{https://www.acm.org/publications/policies/artifact-review-and-badging-current}{
        \includegraphics[height=65pt]{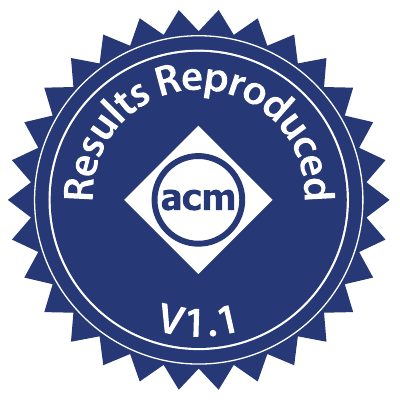}}
      \hspace{80pt}
    }
  }
}
}

\mlsystitlerunning{BLASST\@: Dynamic BLocked Attention Sparsity via Softmax Thresholding}

\begin{document}

\twocolumn[
\mlsystitle{BLASST\@: Dynamic BLocked Attention Sparsity\\ via Softmax Thresholding}



\mlsyssetsymbol{equal}{*}
\mlsyssetsymbol{mark}{\dag}

\begin{mlsysauthorlist}
\mlsysauthor{Jiayi Yuan}{equal,rice}
\mlsysauthor{Cameron Shinn}{equal,ucd}
\mlsysauthor{Kai Xu}{nv}
\mlsysauthor{Jingze Cui}{nv}
\mlsysauthor{George Klimiashvili}{nv}
\mlsysauthor{Guangxuan Xiao}{nv}
\mlsysauthor{Perkz Zheng}{nv}
\mlsysauthor{Bo Li}{nv}
\mlsysauthor{Yuxin Zhou}{nv}
\mlsysauthor{Zhouhai Ye}{nv}
\mlsysauthor{Weijie You}{nv}
\mlsysauthor{Tian Zheng}{nv}
\mlsysauthor{Dominic Brown}{nv}
\mlsysauthor{Pengbo Wang}{nv}
\mlsysauthor{Markus Hoehnerbach}{meta,mark}
\mlsysauthor{Richard Cai}{nv}
\mlsysauthor{Julien Demouth}{nv}
\mlsysauthor{John D. Owens}{ucd}
\mlsysauthor{Xia Hu}{rice}
\mlsysauthor{Song Han}{nv}
\mlsysauthor{Timmy Liu}{nv}
\mlsysauthor{Huizi Mao}{nv}
\end{mlsysauthorlist}

\mlsysaffiliation{rice}{Rice University, Houston, Texas, USA}
\mlsysaffiliation{ucd}{University of California, Davis, California, USA}
\mlsysaffiliation{nv}{NVIDIA, Santa Clara, California, USA}
\mlsysaffiliation{meta}{Meta, Menlo Park, California, USA\@.\ \textsuperscript{\dag}Work completed while at NVIDIA}

\mlsyscorrespondingauthor{Timmy Liu}{jiliu@nvidia.com}
\mlsyscorrespondingauthor{Huizi Mao}{huizim@nvidia.com}

\mlsyskeywords{Machine Learning, MLSys}

\vskip 0.3in

\begin{abstract}
The growing demand for long-context inference capabilities in Large Language Models (LLMs) has intensified the computational and memory bottlenecks inherent to the self-attention mechanism. To address this challenge, we introduce \algname{}, a drop-in, dynamic sparse attention mechanism that accelerates inference by using only a fixed scalar threshold to skip attention blocks. Our method targets practical inference deployment by removing the barriers to adoption present in existing works. As such, \algname{} eliminates training requirements, avoids expensive pre-computation passes, accelerates both prefill and decode across all major attention variants (MHA, GQA, MQA, and MLA), provides optimized support for modern hardware, and easily integrates into existing frameworks. This is achieved by reusing online softmax statistics to identify negligible attention scores, skipping softmax, value block loads, and the subsequent matrix multiplication. We demonstrate the \algname{} algorithm by delivering optimized kernels with negligible latency overhead. Our automated threshold calibration procedure reveals a simple inverse relationship between optimal threshold and context length, meaning we require only a single threshold each for prefill and decode per model. Preserving benchmark accuracy, we demonstrate a 1.52$\times$ speedup for prefill at 71.9\% sparsity and a 1.48$\times$ speedup for decode at 73.2\% sparsity on modern GPUs.

\end{abstract}
] 


\printAffiliationsAndNotice{\mlsysEqualContribution} 

\section{Introduction}

Large Language Models (LLMs) have revolutionized natural language processing, achieving remarkable performance across diverse tasks. However, their practical deployment faces a critical bottleneck: the quadratic computational complexity of the attention mechanism. As applications increasingly demand longer context windows---from processing entire codebases~\cite{roziere2023code} to analyzing lengthy documents~\cite{zeng2025glm} and maintaining extended conversations~\cite{achiam2023gpt}---this bottleneck becomes increasingly severe. Recent models like Deepseek-R1~\cite{guo2025deepseek} and Qwen3~\cite{yang2025qwen3} support context lengths up to 128K tokens, with some models pushing to 1M tokens~\cite{comanici2025gemini}. Yet processing such long sequences remains computationally prohibitive, with attention computation dominating both latency and me-
\begin{figure}[H]
\centering
\vspace{1em}
\includegraphics[width=0.8\columnwidth]{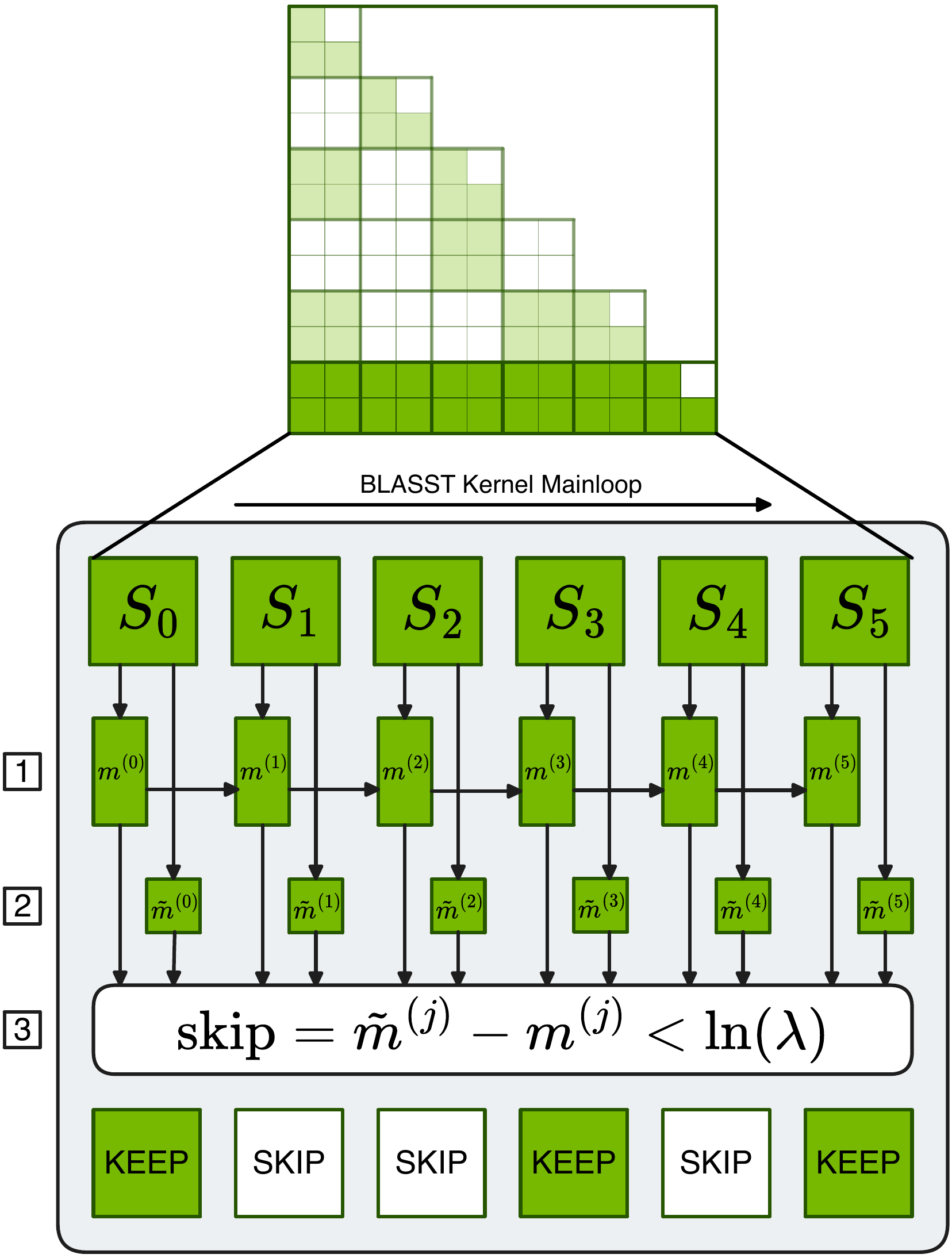}
\vspace{-1em}
\caption{Overview of \algname{}. Blocks along a row of the attention matrix are sequentially processed. We (1) update the running row max ($m^{(j)}$) as in FlashAttention, (2) compute the block max ($\tilde{m}^{(j)}$) for each $S_j$ block ($QK_j^\top$), and (3) skip subsequent work if the block max is lower than the running max by more than the input threshold, $\ln(\lambda)$. Full details can be found in Algorithm~\ref{alg:skip_softmax}.}
\label{fig:teaser}
\end{figure}
mory consumption. For a sequence of length $n$, the attention mechanism requires $O(n^2)$ operations and memory accesses, making real-world deployment of long-context models challenging even with state-of-the-art hardware. While FlashAttention~\cite{dao2022flashattention, zadouri2026flashattention} and its successors have optimized memory bandwidth utilization through tiling and kernel fusion, they still compute the full attention matrix, leaving the fundamental quadratic complexity unaddressed.

\emph{Sparse attention} methods have emerged as a promising solution by computing only a subset of the full attention matrix. While these approaches cleverly determine which attention scores to skip, their added complexity hinders practical use. We identify five key hurdles to their adoption: (1)~Some methods require expensive pre-computation to determine sparsity patterns, often negating their theoretical speedups~\cite{jiang2024minference, xu2025xattention}. (2)~Other methods introduce new layers that require model fine-tuning~\cite{xiao2024duoattention} or training an entirely new architecture~\cite{deepseekai2024deepseekv32}. (3)~Most existing works focus exclusively on either the prefill or decode phase, missing opportunities for end-to-end inference acceleration. (4)~They lack kernel support for newer GPUs, making it unclear if their speedups translate to the characteristics of modern hardware, like Blackwell and Hopper. (5)~These methods often hinder framework integration, requiring intrusive modifications to model architectures or attention interfaces and substantial changes to existing APIs.

To address these hurdles, we present \algname{} (BLocked Attention Sparsity via Softmax Thresholding), a simple yet effective \textbf{training-free} sparse attention method that dynamically prunes negligible attention blocks during \textbf{both prefill and decode} with \textbf{no pre-computation overhead}. Our key insight is that during FlashAttention's block-wise online-softmax, we can identify and skip blocks whose contribution to the final output will be negligible based solely on already-computed information. Specifically, when processing blocks sequentially, we maintain a running maximum of attention scores. As shown in Figure~\ref{fig:teaser}, if a block's local maximum score is significantly smaller than this running maximum (by a threshold $\lambda$), its post-softmax values will be near zero after normalization. We can therefore skip three expensive operations for such blocks: (1)~computing the exponential for softmax, (2)~loading the corresponding value block from HBM, and (3)~multiplication between attention and values. This simple pruning rule requires only a single comparison per block and \textbf{seamlessly integrates into existing attention APIs}, requiring only a single scalar threshold input.

To maximize the practical impact of \algname{}, we provide \textbf{optimized CUDA kernels for Blackwell and Hopper} that implement our sparse attention algorithm. Our kernels are designed with two key goals: (1)~introduce minimal overhead for the block-skipping decision logic by reusing already-computed statistics, and (2)~strategically target the bottleneck resources in each phase---reducing CUDA core and tensor core usage in compute-bound prefill, and reducing memory bandwidth consumption in memory-bound decode. Our prefill and decode kernels are tailored to their distinct computational patterns. Our kernels achieve up to 1.52$\times$ speedup for prefill at 71.9\% sparsity and 1.48$\times$ speedup for decode at 73.2\% sparsity over FlashAttention baselines~\cite{shah2024flashattention, zadouri2026flashattention}, while maintaining numerical stability and supporting the common attention variants (MHA, MQA, GQA, MLA).

Beyond the core algorithm and kernel implementation, we develop two key techniques to enhance \algname{}'s deployment and performance. First, we propose an automated calibration procedure that determines optimal thresholds for any target sparsity level. Our calibration reveals a robust inverse relationship $\lambda = a/L$ between threshold and context length $L$, enabling reliable deployment across diverse scenarios without manual tuning. Second, we explore sparsity-aware training as a natural extension, showing that models can be trained to be inherently more robust to sparse attention patterns. This training approach further pushes the accuracy-sparsity frontier, enabling even higher sparsity levels with minimal loss in accuracy.

Our contributions include:

\begin{enumerate}[nosep, leftmargin=*]
    \item The \algname{} algorithm, a drop-in method with no pre-computation overhead and no proxy scores, achieving minimal accuracy loss.
    \item Automated hyperparameter selection and sparsity-aware training for robust, flexible, and extensible deployment.
    \item Optimized CUDA kernels implementing \algname{} for both prefill and decode, available in TensorRT-LLM\footnote{GPU kernels and inference framework support can be found at https://github.com/NVIDIA/TensorRT-LLM} and FlashInfer.
\end{enumerate}

\section{Related Works}

Effectively exploiting the sparse attention property requires either reducing compute on unimportant interactions or reducing memory footprint (e.g., KV cache) without expensive selection overheads or retraining. Comparing to the following related works, \algname{} addresses both dimensions simultaneously, in a training-free manner. Table~\ref{tab:attention_methods_updated} summarizes the landscape of existing work.

\begin{table}[t]
    \centering
    \caption{Feature comparison of sparse attention methods. \algname{} distinguishes itself as the only method capable of accelerating both prefill and decode phases without requiring training or costly pre-computation steps.}
    \vspace{.5em}
    \label{tab:attention_methods_updated}
    \setlength{\tabcolsep}{0pt}
    \footnotesize
    \begin{tabular*}{\columnwidth}{@{\extracolsep{\fill}}lcccc}
        \toprule
        {Method} & {\makecell{Accelerates\\Prefill}} & {\makecell{Accelerates\\Decode}} & {\makecell{No\\Training}} & {\makecell{No Pre-\\Computation}} \\
        \midrule
        H2O             & \xmark & \cmark & \cmark & \cmark \\
        SnapKV          & \xmark & \cmark & \cmark & \cmark \\
        RocketKV        & \xmark & \cmark & \cmark & \xmark \\
        Quest           & \xmark & \cmark & \cmark & \xmark \\
        DuoAttention    & \cmark & \cmark & \xmark & \cmark \\
        DSA             & \cmark & \cmark & \xmark & \cmark \\
        MInference      & \cmark & \xmark & \cmark & \xmark \\
        SpargeAttention & \cmark & \xmark & \cmark & \xmark \\
        XAttention      & \cmark & \xmark & \cmark & \xmark \\
        \algname{}      & \cmark & \cmark & \cmark & \cmark \\
        \bottomrule
        \vspace{-2em}
    \end{tabular*}
\end{table}

\subsection{Compute-Optimized Sparsity}
Several approaches reduce attention \emph{compute} by selecting important interactions. Static pattern methods like Sparse Transformer~\cite{child2019generating}, LongFormer~\cite{beltagy2020longformer}, and BigBird~\cite{zaheer2020big} reduce complexity through local or block-based attention. Retrieval head-based methods~\cite{wu2024retrieval,xiao2024duoattention} accelerate model decoding by focusing compute on crucial retrieval heads. Dynamic sparsity methods like MInference~\cite{jiang2024minference} use pre-computed importance scores, XAttention~\cite{xu2025xattention} ranks anti-diagonal blocks, and FlexPrefill~\cite{lai2025flexprefill} offers compiler-supported, flexible block patterns; while effective for prefill, their pre-computation and scheduling overheads can limit realized speedups. Training-aided sparsity such as SeerAttention~\cite{gao2025seerattention} induces high sparsity via (pre)training gates, improving efficiency but adding training cost and showing mixed downstream model performance. {FLASH-D}~\cite{alexandridis2025flashd} leverages the mathematical properties of online softmax in a similar way as \algname{}, but to improve numerical stability and parallelism on custom hardware accelerators.

SpargeAttention~\cite{zhang2025spargeattn} has the most similar design to \algname{}. We differ in three key aspects: (1)~\algname{} optimizes both prefill and decode with specialized kernels, while SpargeAttention targets prefill only; (2)~we make skip decisions directly using already-computed statistics with zero overhead, while SpargeAttention uses a separate prediction step; (3)~our decode kernel skips Value loading from HBM, addressing memory-bound bottlenecks on top of compute savings. In addition, we provide automated calibration and sparsity-aware training.

\subsection{Memory-Optimized Sparsity}
Token/KV sparsity focuses on reducing \emph{memory} footprint and decode-time cost. H2O~\cite{zhang2023h2o}, TOVA~\cite{oren2024transformers}, and InfLLM~\cite{xiao2024infllm} discard tokens based on query patterns. StreamingLLM~\cite{xiao2023efficient} retains initial and recent tokens for consistent latency and memory usage. Quest~\cite{tang2024quest} prunes tokens conditioned on the current query, Rectified Sparse Attention~\cite{sun2025rectified} adaptively selects tokens to maintain accuracy at high sparsity, RocketKV~\cite{behnam2025rocketkv} compresses the KV cache with selective eviction, and recent KV compression for hyper-scaling~\cite{lancucki2025inference} further extends effective context; TidalDecode~\cite{yang2024tidaldecode} stabilizes decode efficiency with position-persistent patterns. We further distinguish reasoning-oriented compression methods such as RPC~\cite{song2025reasoning}, which prioritize preserving reasoning-critical information under memory constraints; non-eviction methods such as Loki~\cite{singhania2024loki}, which avoid explicit KV eviction while reducing effective memory/computation overhead. In general, these methods reduce memory accesses in the decode phase, whereas \algname{} reduces compute and memory accesses in both prefill and decode while remaining training-free.

\subsection{New Attention Variants}
Beyond the above methods, alternative mechanisms include Sliding Window Attention~\cite{beltagy2020longformer}, Linear or Gated Attention~\cite{qiu2025gated}, and State-Space Models (SSM)~\cite{gu2023mamba}. Native Sparse Attention (NSA)~\cite{yuan2025native} and DeepSeek Sparse Attention (DSA)~\cite{deepseekai2024deepseekv32}, while effective in some regimes, often require architectural changes and training. By contrast, \algname{} is a training-free method that accelerates both prefill and decode without proxy scores or complex pre-computation, integrating seamlessly with FlashAttention implementations.

\section{Methodology}

\subsection{Pruning Attention with Running Maximums}

The core insight of \algname{} lies in the observation that during the computation of attention scores in FlashAttention, many blocks contribute negligibly to the final output after softmax normalization. Our method identifies and skips these blocks dynamically during the forward pass, without requiring pre-computation or proxy scores.

\subsubsection{Key Insight}
In the standard attention mechanism, the softmax operation computes:
\begin{equation}
\text{Attention}(Q, K, V) = \text{softmax}\left(\frac{QK^\top}{\sqrt{d_k}}\right)V
\end{equation}

During FlashAttention's block-wise computation, we maintain a running maximum $m_i^{(j)}$ across blocks. If a block's local maximum $\tilde{m}_i^{(j)}$ is significantly smaller than the current running maximum, i.e., $\tilde{m}_i^{(j)} - m_i^{(j)} < \ln(\lambda)$ for some threshold $\lambda$, then after exponentiation:
\begin{equation}
\exp(\tilde{m}_i^{(j)} - m_i^{(j)}) < \lambda \approx 0
\label{eq:main_formula}
\end{equation}

Since the maximum value is bounded by $\lambda$, the block's contribution to the final attention output will be negligible, allowing us to skip its computation entirely.

Intuitively, this criterion follows a three-step approximation. First, the ideal importance of each score $S_{ij}$ is its value relative to the (unknown) global maximum. Second, computing the true maximum on-the-fly is too expensive, so we use the running maximum as a tractable proxy and compare $S_{ij}$ against it. Third, to enable an efficient block-level decision inside the kernel, we replace token-level $S_{ij}$ with the block-local maximum, which yields the inexpensive condition $(\text{block\_max} - \text{running\_max}) < \ln(\lambda)$.

\subsubsection{Algorithm Design}
Algorithm~\ref{alg:skip_softmax} presents our modified FlashAttention forward pass, where the sequence is tiled into $T_r$ query blocks and $T_c$ KV blocks of size $B_c$ each. The key modification is the introduction of a dynamic pruning condition that saves both computation and memory bandwidth. \textbf{Where We Save:} When $\tilde{m}_i^{(j)} - m_i^{(j)} < \ln(\lambda)$ (line 7), we skip:
\begin{enumerate}[nosep, leftmargin=*]
    \item \textbf{Compute savings (CUDA cores):} The expensive $\exp(\cdot)$ operations for computing $\tilde{P}_{ij}$ require multiple instructions per element: \texttt{MUFU.EX2} (exponential), \texttt{FMUL} (multiplication), and \texttt{FADD} (addition). We also skip the \texttt{rowsum} reduction operations (\texttt{FADD} instructions) for normalizing attention weights. For a typical block, this saves thousands of CUDA core instructions.
    \item \textbf{Compute savings (Tensor cores)} The matrix multiplication $\tilde{P}_{ij} V_j$. In the prefill phase, where kernels are compute-bound, avoiding these MMA operations provides a substantial speedup. 
    \item \textbf{Memory bandwidth savings:} Loading the Value block $V_j$ from HBM to SRAM\@. This is particularly critical in decode phase, where attention is memory-bound.
\end{enumerate}

\begin{algorithm}[]
   \caption{FlashAttention with \algname{}}
   \label{alg:skip_softmax}
   \algrenewcommand\algorithmicindent{1em}
\begin{algorithmic}[1]
   \Require Query blocks $\{Q_i\}_{i=1}^{T_r}$, Key blocks $\{K_j\}_{j=1}^{T_c}$, Value blocks $\{V_j\}_{j=1}^{T_c}$, threshold $\lambda$
   \Ensure Output blocks $\{O_i\}_{i=1}^{T_r}$
   \For{$i = 1$ to $T_r$}
       \State Initialize $m_i^{(0)} = -\infty$, $O_i^{(0)} = 0$, $l_i^{(0)} = 0$
       \For{$j = 1$ to $T_c$}
           \State Compute $S_{ij} = Q_i K_j^\top$ \Comment{Attention scores}
           \State $\tilde{m}_i^{(j)} = \rowmax(S_{ij})$ \Comment{Local maximum}
           \State $m_i^{(j)} = \max(m_i^{(j-1)}, \tilde{m}_i^{(j)})$ \Comment{Running maximum}
           \If{$\tilde{m}_i^{(j)} - m_i^{(j)} < \ln(\boldsymbol\lambda)$}
               \State \textbf{continue} \Comment{\textbf{\emph{Skip this block}}}
           \EndIf
           \State $\tilde{P}_{ij} = \exp(S_{ij} - m_i^{(j)})$ \Comment{Compute attn.\ weights}
           \State $l_i^{(j)} = e^{m_i^{(j-1)} - m_i^{(j)}} l_i^{(j-1)} + \rowsum(\tilde{P}_{ij})$
           \State $O_i^{(j)} = e^{m_i^{(j-1)} - m_i^{(j)}} O_i^{(j-1)} + \tilde{P}_{ij} V_j$
       \EndFor
       \State $O_i = O_i^{(T_c)} / l_i^{(T_c)}$ \Comment{Final normalization}
   \EndFor \\
   \Return $\{O_i\}_{i=1}^{T_r}$
\end{algorithmic}
\end{algorithm}

Our approach directly reduces the total amount of computation by dynamically identifying and skipping negligible attention blocks during the forward pass. This simple yet effective modification requires minimal changes to the existing FlashAttention implementation while providing significant computational savings.

\subsection{Calibration for Optimal Sparsity}

A critical challenge in deploying \algname{} is selecting the appropriate threshold $\lambda$ that balances sparsity and accuracy. To understand this relationship, we conducted experiments on Llama-3.1-8B across RULER benchmark challenging subsets (NIAH\_MULTI, VT, FWE) with context lengths from 8K to 64K tokens.

\begin{figure}[]
    \centering
    \begin{subfigure}{0.50\linewidth}
    \centering
    \includegraphics[width=\linewidth]{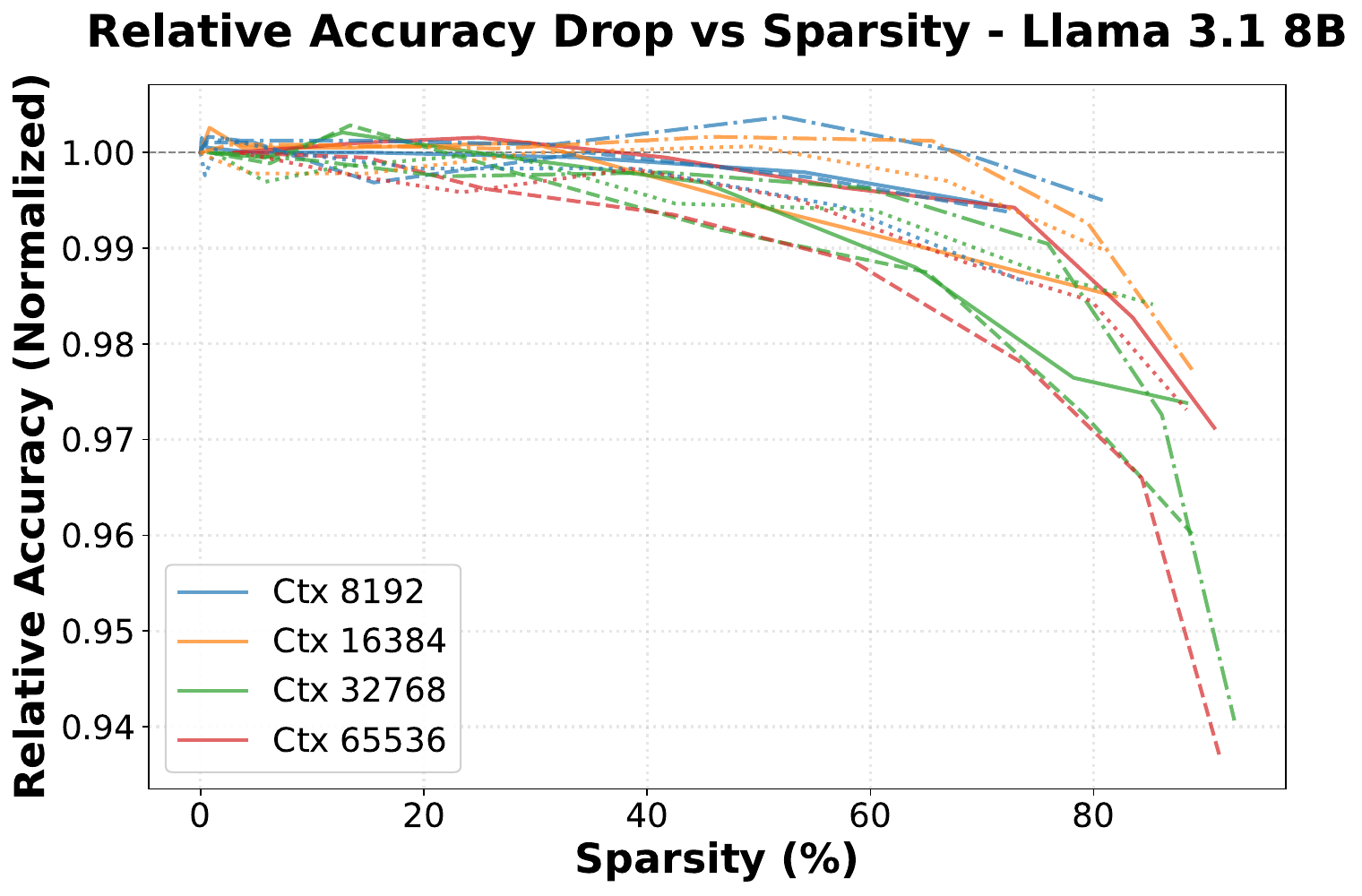}
    \end{subfigure}
    \hfill
    \begin{subfigure}{0.485\linewidth}
    \centering
    \includegraphics[width=\linewidth]{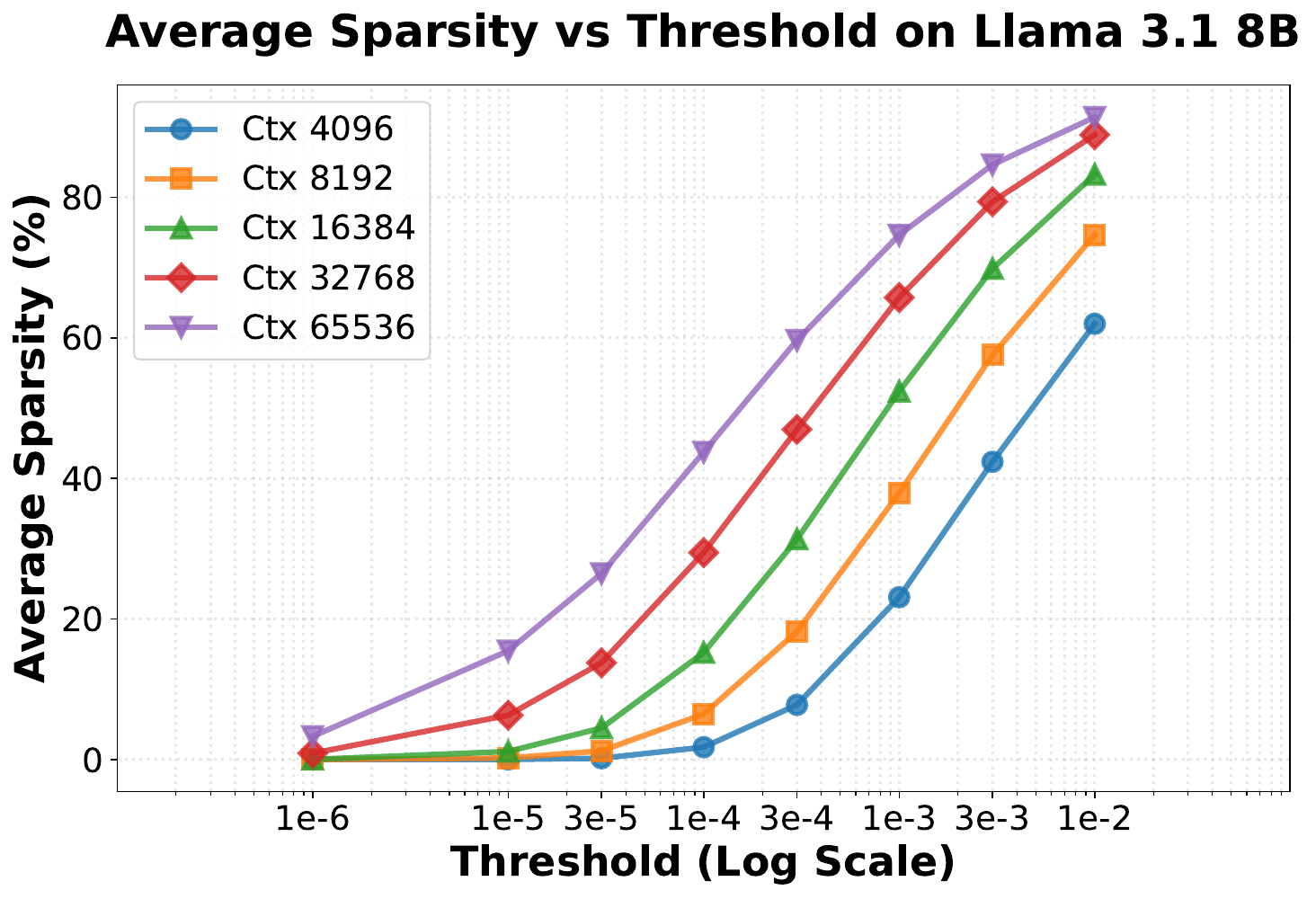}
    \end{subfigure}
    \vspace{-2em}
    \caption{(Left) Relative accuracy drop across different datasets and context lengths shows consistent degradation patterns as observed sparsity increases. All curves are normalized to their initial accuracy. (Right) Relationship between threshold and observed sparsity levels across different sequence lengths, demonstrating the need for threshold calibration to maintain fixed sparsity across varying contexts.}
    \label{fig:inverse}
\end{figure}

\textbf{Sparsity Determines Accuracy.} Figure~\ref{fig:inverse} (left) shows relative accuracy degradation as a function of the observed sparsity level. We normalize each curve by the full attention result for a fair comparison. Remarkably, all curves exhibit consistent degradation patterns: accuracy remains stable up to $\sim$60--70\% sparsity, beyond which accuracy drops sharply. This consistency across diverse tasks and sequence lengths reveals that \textbf{accuracy degradation is primarily determined by the sparsity ratio itself}, not the type of data set or sequence length.

\textbf{Threshold Calibration is Essential.} For models to achieve consistent accuracy, we must maintain a \textbf{fixed sparsity ratio} rather than a fixed threshold. However, Figure~\ref{fig:inverse} (right) shows that achieving 75\% sparsity requires $\lambda \approx \num{1e-4}$ for 8K contexts but only $\num{1e-5}$ for 64K contexts. This necessitates adaptive calibration. Importantly, by targeting fixed sparsity through calibration, users can control and foresee the computational speedup, since accuracy gains scale predictably with the observed sparsity level.

Through empirical analysis, we find that the optimal threshold follows an \textbf{inversely proportional} relationship with context length $L$:
\begin{equation}
\lambda = \frac{a}{L}
\label{eq:threshold_scaling}
\end{equation}
where $a$ is a model-specific scale factor that depends on the target sparsity level. This inverse relationship has theoretical grounding: since attention scores are row-normalized to sum to 1, longer sequences have lower average scores per token, requiring proportionally smaller thresholds. Without calibration, fixed thresholds would cause vastly different sparsity levels across sequence lengths.

\begin{algorithm}[t]
  \caption{\algname{} Calibration}
  \label{alg:calibration}
  \begin{algorithmic}[1]
  \Require Calibration dataset $\mathcal{D} = \{(x_i, L_i)\}_{i=1}^{N}$, threshold set $\Lambda$, sparsity bounds $s_{\min}$, $s_{\max}$
  \Ensure Calibration parameters $\alpha$, $\beta$
  \State Initialize data points $\mathcal{P} = \emptyset$
  \For{each sample $(x_i, L_i)$ in $\mathcal{D}$}
      \For{each $\lambda_j \in \Lambda$}
          \State $s_{ij} \gets \mathrm{MeasureSparsity}(\lambda_j, x_i)$
          \If{$s_{\min} \leq s_{ij} \leq s_{\max}$} \Comment{Filter unreliable extremes}
              \State Add $(\lambda_j \cdot L_i,\; s_{ij})$ to $\mathcal{P}$
          \EndIf
      \EndFor
  \EndFor
  \State Fit exponential model $\lambda \cdot L = \alpha \cdot \exp(\beta \cdot s)$ using $\mathcal{P}$
  \State \Return parameters $\alpha$, $\beta$
  \end{algorithmic}
  \end{algorithm}

To determine $a$ for any target sparsity, we propose the calibration procedure detailed in Algorithm~\ref{alg:calibration}. For each calibration sample $x_i$ of length $L_i$ and each candidate threshold $\lambda_j$, we measure the achieved sparsity $s_{ij}$ and record the scale factor $\lambda_j \cdot L_i$. Since sparsity for all thresholds can be computed from the same attention scores, the entire calibration requires only a single forward pass over $\mathcal{D}$. We then fit an exponential model $\lambda \cdot L = \alpha \cdot \exp(\beta \cdot s)$ to all collected data points. The exponential form reflects the heavy-tailed distribution of attention scores: small increases in threshold prune many low-scoring blocks, while further increases yield diminishing returns as only high-scoring blocks remain. In inference with target sparsity $S$, the threshold is $\lambda = \alpha \cdot \exp(\beta \cdot S) / L$, preserving the inverse relationship Eq.~\eqref{eq:threshold_scaling} while allowing the target sparsity to be adjusted at runtime without recalibration.

More importantly, by targeting fixed sparsity levels, our calibration ensures predictable computational speedup across different context lengths. This is a crucial property for production deployment where consistent performance is required.
We provide additional cross-dataset evidence in Appendix~\ref{sec:appendix_additional} (Table~\ref{tab:dataset_metrics}), showing that the calibrated parameter $a$ yields stable sparsity across diverse tasks without task-specific retuning.

\subsection{Extensibility to Attention Variants}

Because \algname{} depends only on tiled online softmax, it is inherently compatible with many existing dense attention variants. MLA~\cite{deepseekai2024deepseekv32}, for instance, still employs online softmax to compute attention scores within its latent space. Although MLA decoding shifts the bottleneck towards a compute-bound regime, \algname{} remains effective because it eliminates both computation and memory accesses, providing benefits regardless of the primary hardware bottleneck.

\subsection{Sparsity-Aware Training}
\label{sec:sparsity_training}

While \algname{} is primarily designed as a training-free inference optimization, we explore sparsity-aware training as a simple extension to further improve the accuracy-sparsity trade-off. The motivation is straightforward: if models learn to concentrate important information in high-scoring attention blocks during training, they should maintain higher accuracy when those blocks are pruned during inference.

Our method is simple: during fine-tuning, we apply \algname{} in the forward pass to skip negligible attention blocks based on the threshold criterion. In the backward pass, skipped blocks naturally receive no gradients since they were not computed in the forward pass. This encourages the model to adapt its attention patterns to be more compatible with sparsity, concentrating important information in blocks that pass the threshold test. This approach requires no architectural changes or auxiliary losses---it is simply training with the same sparse attention that will be used at inference time.

\section{Kernel Design}

\begin{figure*}[t]
    \centering
    \begin{subfigure}{.9\linewidth}
        \centering
        \caption{Normal FlashAttention-4 prefill pipeline schedule.}
        \includegraphics[width=\linewidth]{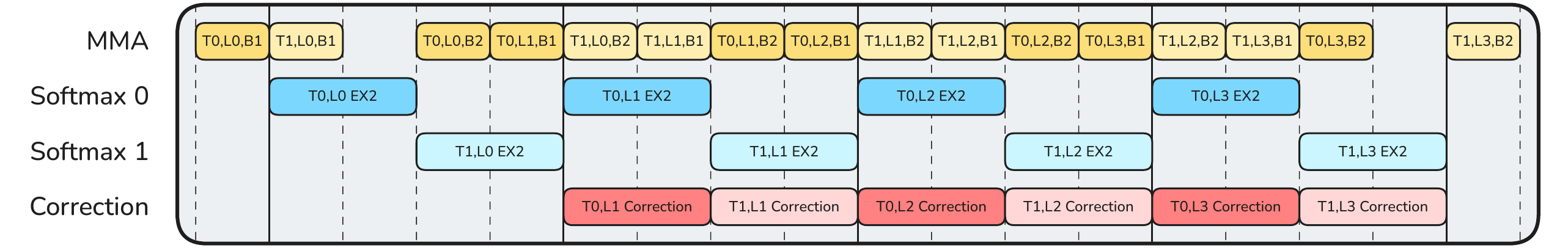}
        \label{fig:prefill_pipeline_fa}
    \end{subfigure}

    \vspace{-1.5em}

    \begin{subfigure}{.9\linewidth}
        \centering
        \caption{\algname{} prefill pipeline schedule with T0 and T1 both skipping loops 1 and 3.}
        \includegraphics[width=\linewidth]{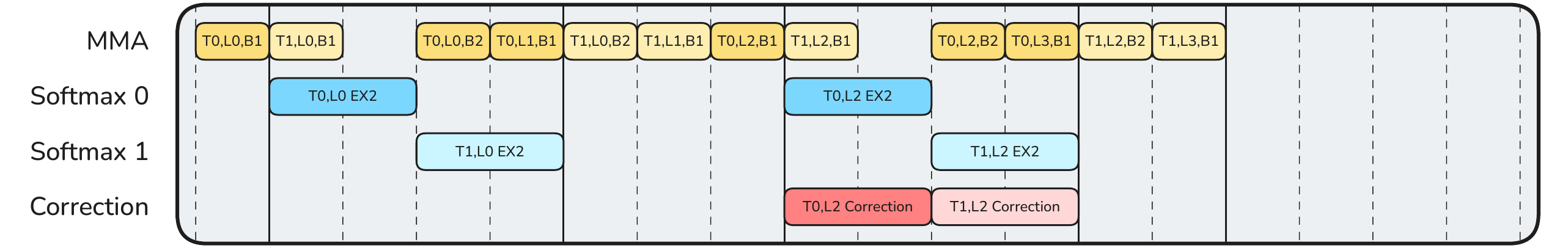}
        \label{fig:prefill_pipeline_blasst_same_1_3}
    \end{subfigure}

    \vspace{-2em}
    \caption{Prefill pipeline schedules for FlashAttention and \algname{} at 50\% sparsity across 4 loop iterations (L0--L3). Rows are separated based on warp/warpgroup specializations. Darker and lighter hues correspond to ops for different tile rows (T0/T1). The MMA warp's BMM1 and BMM2 ops are indicated with B1 and B2. The softmax warpgroups are primarily bottlenecked by exponentiation (EX2), but they also perform the skip check, row sum and softmax scaling (not shown). Mainloop iterations are enclosed by solid lines.}

    \label{fig:prefill_pipeline}
\end{figure*}

The \algname{} kernels were designed with two primary goals: (1)~minimal changes to existing FlashAttention kernel interfaces and implementation structure, and (2)~minimal overhead for block skipping decision logic. Our key insight is to reuse statistics computed during the standard FlashAttention algorithm---specifically, the local maximum and running maximum values maintained in every thread during online softmax. Our optimizations are specific to \algname{} and cannot be applied to typical FlashAttention kernels.

\textbf{Skip Decision Implementation.} The decision process (line 7 in Algorithm~\ref{alg:skip_softmax}) requires only a few additional instructions per block: (1)~setting a predicate per thread based on the threshold comparison, (2)~issuing a \texttt{VOTE} instruction to determine if all threads within a warp agree to skip, and (3)~a single \texttt{ATOMIC} instruction to shared memory issued by one thread per warp to coordinate the block-level decision across the softmax warpgroup. We carefully design the kernel such that the decision-making instructions are hidden behind existing operations, adding negligible latency overhead.

Since prefill and decode phases have fundamentally different performance characteristics, we implement specialized optimizations for each.

\subsection{Prefill Kernel: Compute-Bound Optimization}

Prefill kernels are typically compute-bound, bottlenecked by CUDA core (softmax) and tensor core (matrix multiplication) throughput rather than memory bandwidth. Therefore, our prefill kernel is designed to skip both softmax computation and MMA operations (attention-value multiplication) for pruned blocks.

Figure~\ref{fig:prefill_pipeline} illustrates our changes to the pipeline schedule for the \algname{} prefill kernel, which is optimized for compute-bound scenarios by overlapping different compute tasks. The pipeline schedules operations across Tensor Cores (math warp/matrix multiplication) and CUDA cores (softmax and correction logic). Figure~\ref{fig:prefill_pipeline_blasst_same_1_3} shows that even as all $Q K^\top$ (BMM1) operations are computed, the kernel dynamically skips compute-heavy softmax and attention-value multiplication (BMM2) for blocks identified as negligible (e.g., loop 1 and loop 3 in Figure~\ref{fig:prefill_pipeline_blasst_same_1_3}). By skipping these compute operations, the kernel frees up execution units, allowing subsequent operations to be scheduled earlier. This compresses the entire schedule, reducing the total runtime from 18 time units in Figure~\ref{fig:prefill_pipeline_fa} to 14 units in Figure~\ref{fig:prefill_pipeline_blasst_same_1_3}.

The Value blocks remain loaded from HBM in the prefill kernel because: (1)~memory bandwidth is not the bottleneck, (2)~the prefetching pipeline benefits from predictable memory access patterns, and (3)~the latency of conditional Value loading would exceed the savings. By focusing on eliminating compute operations, we achieve speedups that scale nearly linearly with sparsity in the compute-bound regime. Our current design prioritizes the common case where prefill is compute-bound on modern GPUs; however, Value loading could be skipped in prefill if future workloads and/or hardware architectures shift to a memory-bandwidth-bound regime.

\begin{figure*}[]
    \centering
    \begin{subfigure}{.8\linewidth}
        \centering
        \caption{Normal FlashAttention-4 decode pipeline schedule.}
        \includegraphics[width=\linewidth]{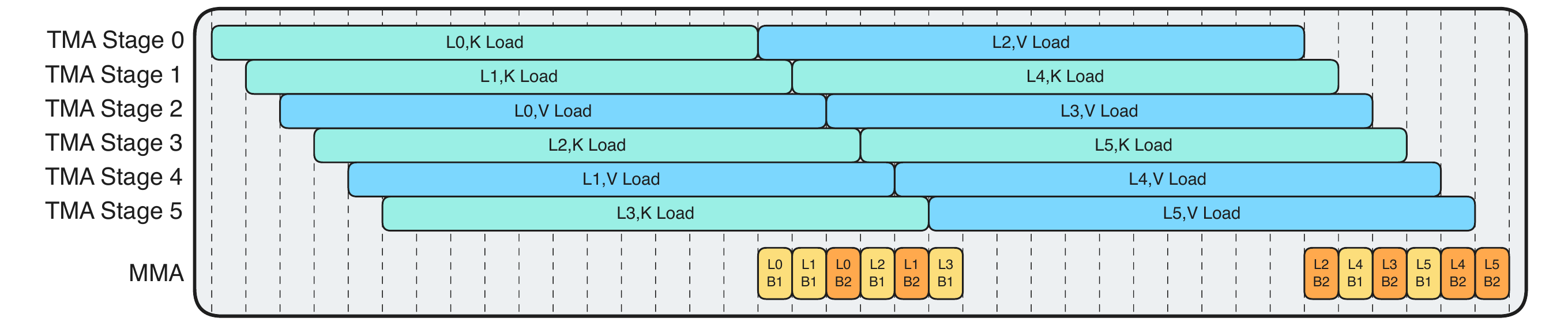}
        \label{fig:decode_pipeline_fa}
    \end{subfigure}

    \vspace{-1.5em}

    \begin{subfigure}{.8\linewidth}
        \centering
        \caption{\algname{} decode pipeline schedule when skipping loops 1, 2, and 4.}
        \includegraphics[width=\linewidth]{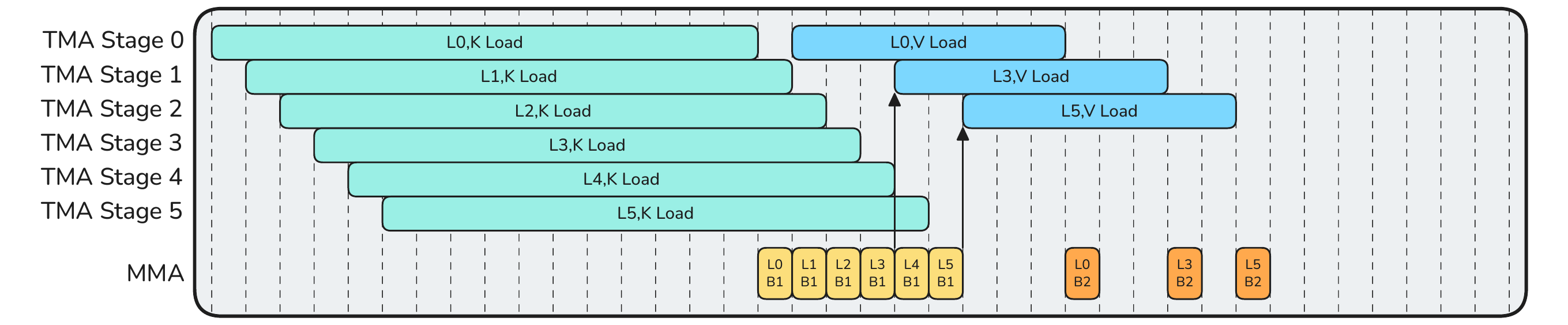}
        \label{fig:decode_pipeline_blasst_1_2_4}
    \end{subfigure}

    \vspace{-2em}
    \caption{Decode pipeline schedules for FlashAttention and \algname{} skipping in loops 1, 2, and 4. We focus on the steady state of the first 6 mainloop iterations (L0--L5). The \algname{} schedule does consecutive K loads since V cannot be pre-fetched until the skip-check is computed after BMM1. V loads in Figure~\ref{fig:decode_pipeline_blasst_1_2_4} finish more quickly because there are fewer simultaneous loads. Arrows indicate scoreboard dependencies from the skip check after BMM1. Note that the MMA warp's BMM1 and BMM2 ops are indicated with B1 and B2.}
    \label{fig:decode_pipeline}
\end{figure*}

\subsection{Decode Kernel: Memory-Bound Optimization}

Decode kernels are typically memory-bound, bottlenecked by the HBM bandwidth required to fetch the KV cache rather than compute, as attention involves only a single Query against all Keys. Our kernel thus focuses on skipping the memory-intensive load of the Value matrix $V_j$ for pruned blocks, directly addressing this HBM bottleneck. This optimization cuts memory traffic proportionally to the sparsity level, while we overlap the threshold and Key operations with the remaining Value loads to achieve a substantial speedup, reflecting the different performance characteristics of decode versus prefill.

A critical challenge in the decode kernel is that in a naive implementation, the value load and subsequent attention-value multiplication (BMM2) would be issued before the query-key multiplication (BMM1) completes and the skip decision can be made. This would result in wasted memory bandwidth loading values that will ultimately be discarded. By conditionally loading blocks of V we can save memory bandwidth; however, this introduces a scoreboard dependency that stops us from issuing consecutive loads ahead of time. The pipeline becomes serialized and can introduce pipeline bubbles.

To address this, we redesign the decode kernel pipeline to use batched load scheduling. As shown in Figure~\ref{fig:decode_pipeline_blasst_1_2_4}, instead of processing blocks end-to-end one at a time, we process multiple consecutive query-key products back-to-back ($K_1^\top Q$, $K_2^\top Q$ \ldots\ $K_B^\top Q$). The tradeoff is that we must maintain $B$ number of shared memory buffers for $S_j$ (from $K_j^\top Q$), but they are relatively small with a query sequence length of 1. This reordering allows us to issue a batch of loads for only $V_j$ tiles that pass the threshold check, removing the possibility of pipeline bubbles. As a result, Figure~\ref{fig:decode_pipeline_fa} takes 38 time units to complete all V loads, whereas Figure~\ref{fig:decode_pipeline_blasst_1_2_4} takes 31 units.

For attention mechanisms like Multi-head Latent Attention (MLA)~\cite{liu2024deepseek} that can be compute-bound even in decode, we also skip softmax operations for pruned blocks, providing further speedup beyond memory savings.

\begin{table*}[t]
\centering
\caption{Performance of \algname{} at different sparsity levels across all models and benchmarks. We evaluate on Llama-3.1-8B and Qwen3-8B across three deployment scenarios: prefill-only optimization (long-context tasks: RULER, LongBench); decode-only optimization (reasoning tasks: MATH500, AIME 2024, GPQA); and combined prefill+decode optimization. Results show minimal accuracy degradation even at $\sim$75\% sparsity, with occasional improvements over the dense baseline.}
\vspace{.5em}
\label{tab:summary_results}
\resizebox{\linewidth}{!}{%
\begin{tabular}{llccccccc}
\toprule
\multirow{2}{*}{\textbf{Model}} & \multirow{2}{*}{\textbf{Target Sparsity}} & \multicolumn{2}{c}{\textbf{Prefill Phase}} & \multicolumn{3}{c}{\textbf{Decode Phase}} & \multicolumn{2}{c}{\textbf{Prefill + Decode Phase}} \\
\cmidrule(lr){3-4} \cmidrule(lr){5-7} \cmidrule(lr){8-9}
& & \textbf{RULER-32K} & \textbf{LongBench} & \textbf{MATH500} & \textbf{AIME2024} & \textbf{GPQA} & \textbf{RULER-32K} & \textbf{LongBench} \\
\midrule
\multirow{3}{*}{Llama-3.1-8B}
& Dense & 92.33 & 31.40 & 73.40 & 46.66 & 46.71 & 92.33 & 31.40 \\
& 50\% & 91.81 & 31.80 & 73.71 & 46.15 & 46.31 & 91.79 & 32.40 \\
& 75\% & 91.67 & 31.80 & 73.89 & 46.01 & 45.95 & 91.67 & 31.80 \\
\midrule
\multirow{3}{*}{Qwen3-8B}
& Dense & 91.90 & 33.60 & 95.87 & 75.00 & 61.21 & 91.90 & 33.60 \\
& 50\% & 92.08 & 35.10 & 96.23 & 76.50 & 61.56 & 92.07 & 33.30 \\
& 75\% & 92.11 & 34.40 & 96.07 & 75.33 & 61.51 & 91.74 & 33.10 \\
\bottomrule
\end{tabular}%
}
\end{table*}

\section{Experiments}

\subsection{Experimental Setup}

\textbf{Models.} We evaluate \algname{} on state-of-the-art language models to demonstrate its effectiveness across different architectures. Our evaluation focuses on two 8B parameter models---Llama-3.1-8B-Instruct and Qwen3-8B-Instruct---both supporting context lengths up to 128K tokens. For long-generation reasoning tasks, we use Llama-3.1-8B-Instruct distilled from DeepSeek-R1~\cite{guo2025deepseek}, which provides enhanced reasoning capabilities while maintaining compatibility with our sparse attention approach.

\textbf{Baselines.} We compare \algname{} against dense attention and SOTA sparse attention methods. For prefill optimization, we compare against MInference~\cite{jiang2024minference}, FlexPrefill~\cite{lai2025flexprefill}, and XAttention~\cite{xu2025xattention}. For decode optimization, we evaluate against Quest~\cite{tang2024quest}, RocketKV~\cite{behnam2025rocketkv}. For each baseline, we adopt its best-performing configuration as reported in its respective paper to ensure fair comparisons.

\textbf{Datasets.} We evaluate on two categories of benchmarks: (1)~\textbf{Long-context tasks}: RULER~\cite{hsieh2024ruler} (synthetic retrieval and reasoning from 4K-128K tokens) and LongBench v2~\cite{bai2024longbench} (real-world QA, summarization, and code completion). (2)~\textbf{Reasoning tasks}: MATH500 (mathematical problem solving), AIME 2024 (advanced mathematics), GPQA (graduate-level science), and LiveCodeBench (code generation). These reasoning benchmarks test whether sparse attention preserves complex multi-step reasoning capabilities. We use the NVIDIA NeMo-Skills framework\footnote{https://github.com/NVIDIA-NeMo/Skills} for standardized evaluation of reasoning tasks.

\textbf{Implementation Details.} We implement \algname{} as optimized CUDA kernels integrated into TensorRT-LLM and FlashInfer~\cite{ye2025flashinfer}. For calibration (Algorithm~\ref{alg:calibration}), we sample approximately 1000 sequences from the RULER dataset across different context lengths (4K, 8K, 16K, 32K, 64K) to fit the calibration parameters $\alpha$ and $\beta$ for the threshold relationship $\lambda = \alpha \cdot \exp(\beta \cdot S) / L$. For sparsity-aware training, we adopt the curriculum training approach from ProLong~\cite{gao2024train}, applying \algname{} during the finetuning phase with a fixed sparsity threshold.

For evaluation, we use different sampling strategies depending on the task type. For long-context benchmarks (RULER and LongBench), we use greedy decoding with temperature=0 and perform a single run per example to ensure deterministic and reproducible results. For reasoning tasks that benefit from sampling diversity, we use temperature=0.6 and top-p=0.95. Specifically, we generate 10 samples per problem for MATH500, GPQA, and LiveCodeBench, and 20 samples per problem for AIME 2024 due to its greater difficulty. For these reasoning tasks, we report the best-of-N performance where the final answer is selected using majority voting or self-consistency.

\begin{table*}[]
\centering
\caption{Prefill phase comparison on Llama-3.1-8B-Instruct across RULER and LongBench. Best (non-dense) score in each column is denoted in bold. Targeting 50\% sparsity, \algname{} achieves the best accuracy among all sparse attention methods, closely matching dense attention in addition to being the easiest to use.}
\vspace{.5em}
\label{tab:prefill_comparison}
\resizebox{.9\linewidth}{!}{%
\begin{tabular}{lcccccccccccc}
\toprule
\multirow{2}{*}{\textbf{Method}} & \multicolumn{6}{c}{\textbf{RULER}} & \multicolumn{6}{c}{\textbf{LongBench}} \\
\cmidrule(lr){2-7} \cmidrule(lr){8-13}
& \textbf{4K} & \textbf{8K} & \textbf{16K} & \textbf{32K} & \textbf{64K} & \textbf{Average} & \textbf{Easy} & \textbf{Hard} & \textbf{Short} & \textbf{Medium} & \textbf{Long} & \textbf{Overall} \\
\midrule
Dense Attention & 96.16 & 95.07 & 94.80 & 92.33 & 87.69 & 93.21 & 29.7 & 32.5 & 38.3 & 28.8 & 25.0 & 31.4 \\
FlexPrefill & 95.99 & 93.67 & 92.73 & 88.14 & 81.14 & 87.72 & 28.8 & 23.8 & 24.4 & 26.5 & 26.2 & 25.7 \\
MInference & \textbf{96.54} & 94.06 & 91.37 & 85.79 & 83.03 & 84.15 & 28.6 & \textbf{32.8} & 36.7 & 30.2 & 24.1 & 31.2 \\
XAttention & 96.37 & 94.47 & 94.48 & \textbf{91.91} & 85.01 & 92.44 & 29.2 & 31.5 & \textbf{38.3} & 26.0 & \textbf{26.9} & 30.6 \\
\midrule
\algname{} ($\sim$50\%) & 96.17 & \textbf{94.70} & \textbf{94.61} & 91.81 & \textbf{87.06} & \textbf{92.87} & \textbf{30.7} & 32.5 & \textbf{38.3} & \textbf{29.8} & 25.0 & \textbf{31.8} \\
\bottomrule
\end{tabular}%
}
\end{table*}

\begin{table*}[]
\centering
\caption{Decode phase comparison on Qwen3-8B across diverse reasoning and generation tasks. Best (non-dense) score in each column is denoted in bold. Targeting 50\% sparsity, \algname{} matches or exceeds dense baseline on all benchmarks, including mathematical reasoning (MATH500, AIME 2024), graduate-level science (GPQA), and code generation (LiveCodeBench), while maintaining long-context performance (RULER, LongBench).}
\vspace{.5em}
\label{tab:decode_comparison}
\resizebox{.8\linewidth}{!}{%
\begin{tabular}{lccccccc}
\toprule
\textbf{Method} & \textbf{RULER-32K} & \textbf{LongBench} & \textbf{MATH500} & \textbf{AIME 2024} & \textbf{LiveCodeBench} & \textbf{GPQA} & \textbf{Average} \\
\midrule
Dense Attention & 91.90 & 33.60 & 95.87 & 75.00 & 53.83 & 61.21 & 68.57 \\
Quest & 56.23 & 30.30 & 94.18 & 71.50 & 52.17 & 60.12 & 60.75 \\
RocketKV & 87.89 & 30.60 & 95.88 & 73.54 & 53.10 & 60.50 & 66.91 \\
\midrule
\algname{} $\sim$50\% & \textbf{91.55} & \textbf{33.90} & \textbf{96.23} & \textbf{76.50} & \textbf{54.15} & \textbf{61.51} & \textbf{68.97} \\
\bottomrule
\end{tabular}%
}
\end{table*}

\subsection{Main Results}

\textbf{Overall Performance.} Table~\ref{tab:summary_results} presents the accuracy results of \algname{} at 50\% and 75\% target sparsity levels on Llama-3.1-8B and Qwen3-8B across a diverse set of language benchmarks. We also evaluate larger model variants on LongBench and NIAH (Table~\ref{tab:large_model_longbench} and Table~\ref{tab:large_model_niah} in Appendix~\ref{sec:appendix_additional}), and evaluate \algname{} on DeepSeek-R1 to demonstrate compatibility with the MLA attention mechanism (Table~\ref{tab:deepseek_accuracy} in Appendix~\ref{sec:appendix_additional}). Remarkably, \algname{} not only maintains accuracy with minimal degradation but occasionally \emph{outperforms} the dense baseline. For example, on Qwen3-8B, we observe improvements on MATH500 (96.23 vs.\ 95.87) and AIME 2024 (76.50 vs 75.00) at 50\% sparsity. This counterintuitive result can be attributed to two factors. First, in long-context tasks where information is inherently sparse, pruning low-attention blocks forces the model to concentrate probability mass on the most relevant tokens, effectively acting as implicit denoising. Second, for long-generation reasoning tasks, some intermediate reasoning steps or tokens may be redundant or even detrimental~\cite{sui2025stop}; by skipping blocks with negligible attention scores, we filter out such distractions, allowing the model to focus on essential reasoning chains. These results show that \algname{} is not only computationally efficient, but also improves response quality in certain scenarios.

\textbf{Prefill Phase Comparison.} Table~\ref{tab:prefill_comparison} compares \algname{} against state-of-the-art prefill-optimized sparse attention methods on Llama-3.1-8B. Across RULER (4K--64K context lengths) and LongBench, \algname{} achieves the best overall accuracy (92.87 RULER average, 31.8 LongBench) among all sparse methods, closely matching dense attention (93.21, 31.4). In particular, \algname{} significantly outperforms MInference (84.15 RULER) and FlexPrefill (87.72 RULER), demonstrating the effectiveness of our threshold-based pruning over proxy-based importance estimation.

\textbf{Decode Phase Comparison.} Table~\ref{tab:decode_comparison} evaluates \algname{} on Qwen3-8B across reasoning-intensive tasks. Targeting 50\% sparsity, \algname{} matches or exceeds dense baseline performance on all benchmarks, while maintaining long-context capabilities. We note that all existing methods employ different optimization strategies and target different deployment scenarios, making direct comparison challenging. We include Quest and RocketKV as reference points to contextualize \algname{}'s performance. For instance, RocketKV shows 87.89 RULER and 30.60 LongBench scores, illustrating the trade-offs between aggressive KV cache compression and \algname{}'s preservation of critical attention patterns.

\begin{table}[t]
\centering
\caption{BLASST speedup over dense baseline on Blackwell and Hopper GPUs. Sparsity values reported are achieved sparsity levels, obtained by varying the threshold~($\lambda$). Prefill: batch size 1, 64K sequence length. B200 decode: batch size 148, 32K sequence length. H200 decode: batch size 128, 16K sequence length.}
\vspace{.5em}
\label{tab:combined_speedup}
\small 
\setlength{\tabcolsep}{8pt}
\begin{tabular}{cc|cc}
\toprule
\multicolumn{2}{c|}{\textbf{Blackwell (B200)}} & \multicolumn{2}{c}{\textbf{Hopper (H200)}} \\
\textbf{Sparsity} & \textbf{Speedup} & \textbf{Sparsity} & \textbf{Speedup} \\
\midrule
\multicolumn{4}{c}{\textit{Prefill Phase}} \\
\midrule
0.0\%  & 1.00$\times$ & 0.0\%  & 1.00$\times$ \\
38.9\% & 1.25$\times$ & 23.8\% & 1.08$\times$ \\
49.2\% & 1.33$\times$ & 49.2\% & 1.27$\times$ \\
63.0\% & 1.43$\times$ & 57.3\% & 1.35$\times$ \\
71.9\% & 1.52$\times$ & 71.0\% & 1.52$\times$ \\
80.8\% & 1.61$\times$ & 79.5\% & 1.64$\times$ \\
88.9\% & 1.71$\times$ & 88.5\% & 1.78$\times$ \\
94.2\% & 1.77$\times$ & 92.0\% & 1.84$\times$ \\
\midrule
\multicolumn{4}{c}{\textit{Decode Phase}} \\
\midrule
0.0\%  & 0.98$\times$ & 0.0\%  & 0.96$\times$ \\
36.9\% & 1.18$\times$ & 23.8\% & 1.08$\times$ \\
46.7\% & 1.25$\times$ & 43.7\% & 1.20$\times$ \\
61.2\% & 1.34$\times$ & 59.4\% & 1.31$\times$ \\
73.2\% & 1.48$\times$ & 70.5\% & 1.40$\times$ \\
82.6\% & 1.64$\times$ & 78.4\% & 1.47$\times$ \\
87.0\% & 1.71$\times$ & 87.5\% & 1.56$\times$ \\
92.0\% & 1.79$\times$ & ---    & ---          \\
\bottomrule
\vspace{-2em}
\end{tabular}
\end{table} 

\subsection{GPU Kernel Performance}

We implement and benchmark highly optimized kernels for both Blackwell (B200) and Hopper (H200) GPU architectures, demonstrating that \algname{} achieves substantial real-world speedups. Table~\ref{tab:combined_speedup} shows performance scaling across increasing sparsity levels for both prefill and decode phases. All speedups are measured against FlashAttention-3 BF16 baselines.

\textbf{Key Results.} At near-lossless accuracy ($\sim$50\% target sparsity), we achieve approximately 1.33$\times$ speedup for prefill and 1.25$\times$ speedup for decode on Blackwell. At higher sparsity ($\sim$70\%), the speedup increases to 1.52$\times$ for prefill and 1.48$\times$ for decode. On Hopper, prefill achieves up to 1.52$\times$ speedup at 71.0\% sparsity. These speedups scale predictably with sparsity: higher sparsity yields greater performance gains, allowing users to choose their preferred accuracy-performance trade-off.

Importantly, we observe no significant performance degradation at 0\% sparsity (0.96--1.00$\times$ baseline), verifying the kernels are able to hide the skip check computation behind Tensor Core (prefill) or HBM load (decode) instructions.

\begin{figure}[H]
    \centering
    \includegraphics[width=\linewidth]{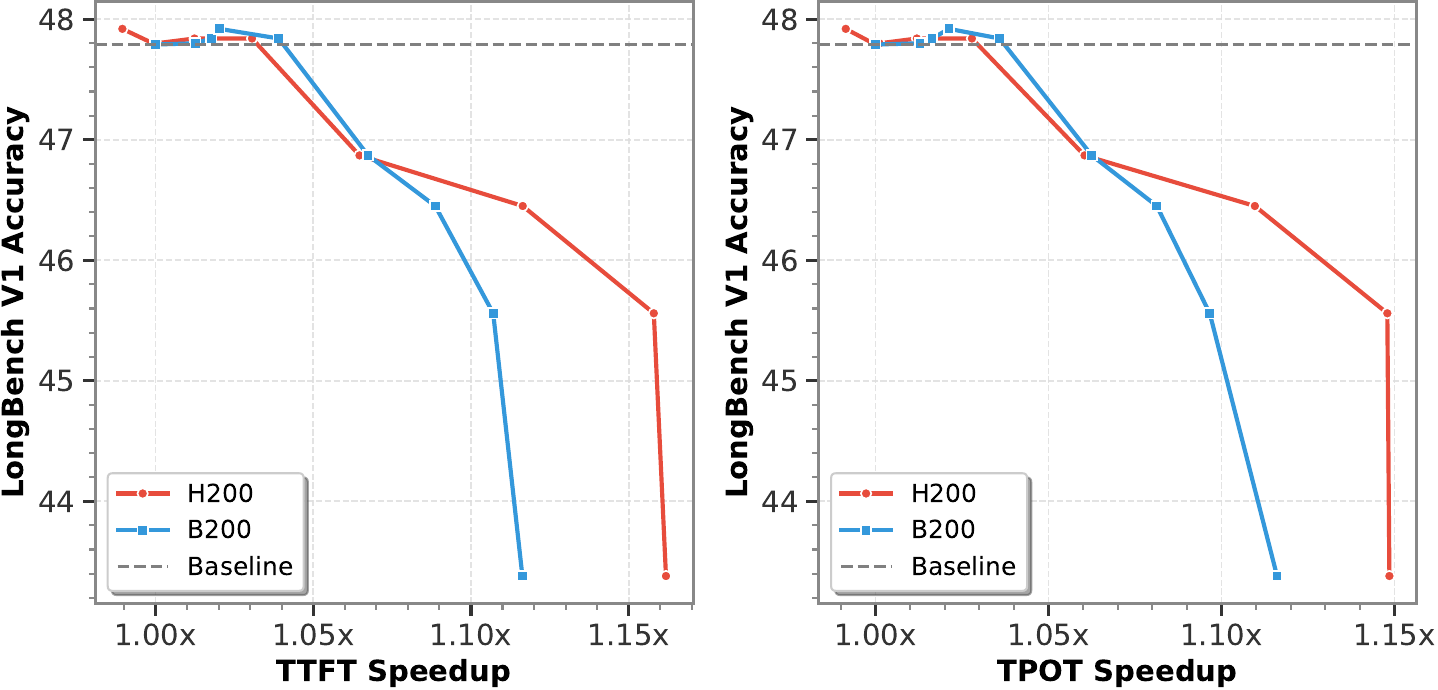}
    \vspace{-2em}
    \caption{\algname{} shows meaningful end-to-end acceleration at medium to long context lengths. Data shows Qwen3-30B-A3B-Instruct-2507, evaluated on LongBench V1 with H200 and B200 GPUs, collected by stepping target sparsity from 0\% to 80\%. Average input sequence length is 10K\@; average output sequence length is 6. Performance is measured in TensorRT-LLM using in-flight batching with concurrency 64, meaning decoding requests may be piggybacked with prefilling requests.}
    \vspace{-2em}
    \label{fig:accuracy_vs_speedup}
\end{figure}

For a full picture, the accuracy-performance tradeoffs in Figure~\ref{fig:accuracy_vs_speedup} show how effective \algname{} is in an inference serving environment. We see a $1.1\times$ speedup in TTFT and TPOT with only a marginal drop in LongBench V1 accuracy.

\subsection{Calibration Results}

A key motivation for our calibration approach is that fixed thresholds produce inconsistent sparsity across different context lengths, making deployment unreliable. Table~\ref{tab:calibration_robust} demonstrates the effectiveness of our calibration method across varying sequence lengths. For a target sparsity of 50\%, the fixed threshold approach makes the observed sparsity highly unstable, ranging from 23\% at 4K to 75\% at 64K, making it impractical for production deployment. In contrast, our calibrated $\lambda = a/L$ approach maintains sparsity within a tight range with an average error of only 1.2\% from the target. Similar improvements are observed at 70\% target sparsity. These results confirm that our calibration enables reliable, predictable sparsity control across diverse sequence lengths without manual tuning.

\begin{table}[t]
\centering
\vspace{-.5em}
\caption{Sparsity stability across context lengths: calibrated vs.\ fixed threshold on Llama-3.1-8B. Our calibration method maintains consistent sparsity levels across different context lengths, while fixed thresholds produce high variance. Values in parentheses indicate deviation of observed sparsity from the target.}
\vspace{.5em}
\label{tab:calibration_robust}
\resizebox{\columnwidth}{!}{%
\begin{tabular}{lccccc}
\toprule
\textbf{Method} & \textbf{4K} & \textbf{8K} & \textbf{16K} & \textbf{32K} & \textbf{64K} \\
\midrule
\multicolumn{6}{c}{\textit{Target Sparsity: 50\%}} \\
\midrule
\multirow[t]{2}{*}{Fixed $\lambda = \num{1e-3}$} & 23.09 & 37.92 & 52.38 & 65.72 & 74.63\\
& (-26.91) & (-12.08) & (+2.38) & (+15.72) & (+24.63) \\
\addlinespace
\multirow[t]{2}{*}{Calibrated $\lambda = a/L$} & 54.20 & 49.70 & 52.20 & 46.96 & 48.75 \\
& (+4.20) & (-0.30) & (+2.20) & (-3.04) & (-1.25) \\
\midrule
\multicolumn{6}{c}{\textit{Target Sparsity: 70\%}} \\
\midrule
\multirow[t]{2}{*}{Fixed $\lambda = \num{3e-3}$} & 42.35 & 57.54 & 69.83 & 79.36 & 84.63 \\
& (-27.65) & (-12.46) & (-0.17) & (+9.36) & (+14.63)\\
\addlinespace
\multirow[t]{2}{*}{Calibrated $\lambda = a/L$} & 67.99 & 74.65 & 73.64 & 72.54 & 74.63 \\
& (-2.01) & (+4.65) & (+3.64) & (+2.54) & (+4.63) \\
\bottomrule
\end{tabular}%
}
\end{table}

Beyond context-length stability, we also verify that the calibrated parameter $a$ transfers across different task types without heavy retuning (Table~\ref{tab:dataset_metrics} in Appendix~\ref{sec:appendix_additional}).

\subsection{Sparsity-Aware Training Results}

\begin{figure}[t]
    \centering
    \includegraphics[width=.85\columnwidth]{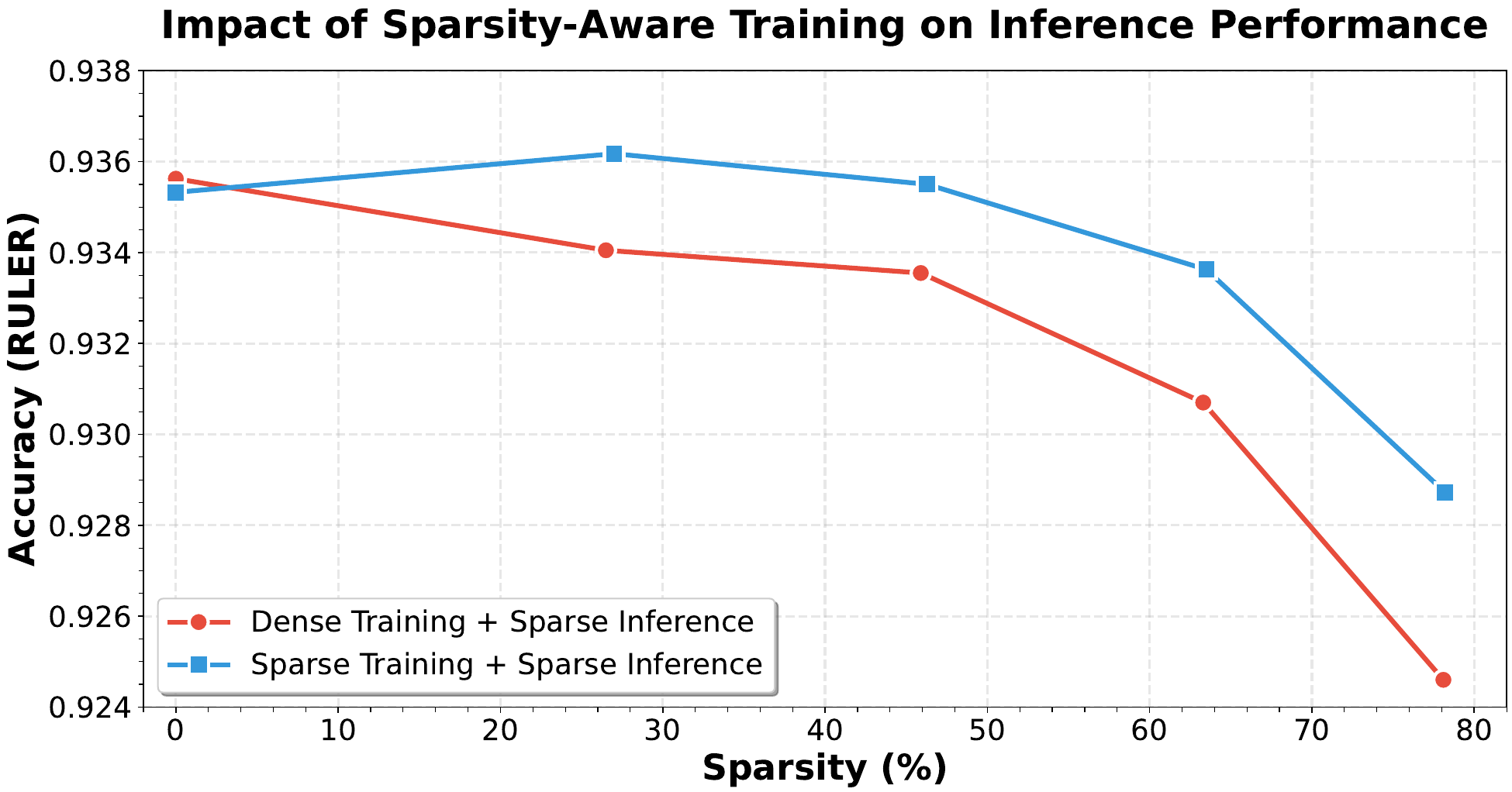}
    \vspace{-1em}
    \caption{Sparsity-aware training pushes the accuracy-sparsity frontier. Models fine-tuned with \algname{} active during training maintain higher accuracy at aggressive (observed) sparsity levels compared to training-free sparsity application. By training with sparse attention, models learn to concentrate information in high-scoring blocks, making them more robust to pruning.}
    \label{fig:sparsity_training}
\end{figure}

Figure~\ref{fig:sparsity_training} demonstrates that sparsity-aware training improves the accuracy-sparsity trade-off on RULER benchmarks. At low sparsity levels, sparse-trained models even slightly outperform the dense baseline, suggesting the model learns more robust attention patterns. In the target sparsity range of 50--75\%, sparse-trained models achieve substantially better accuracy than applying sparsity training-free, reducing accuracy degradation by up to 1.7$\times$. These results confirm that models can be trained to concentrate information in high-scoring attention blocks, making them inherently more compatible with sparse attention patterns and pushing the Pareto frontier of efficient attention.

\subsection{Ablation Studies}

\begin{figure}[t]
    \centering
    \includegraphics[width=\columnwidth]{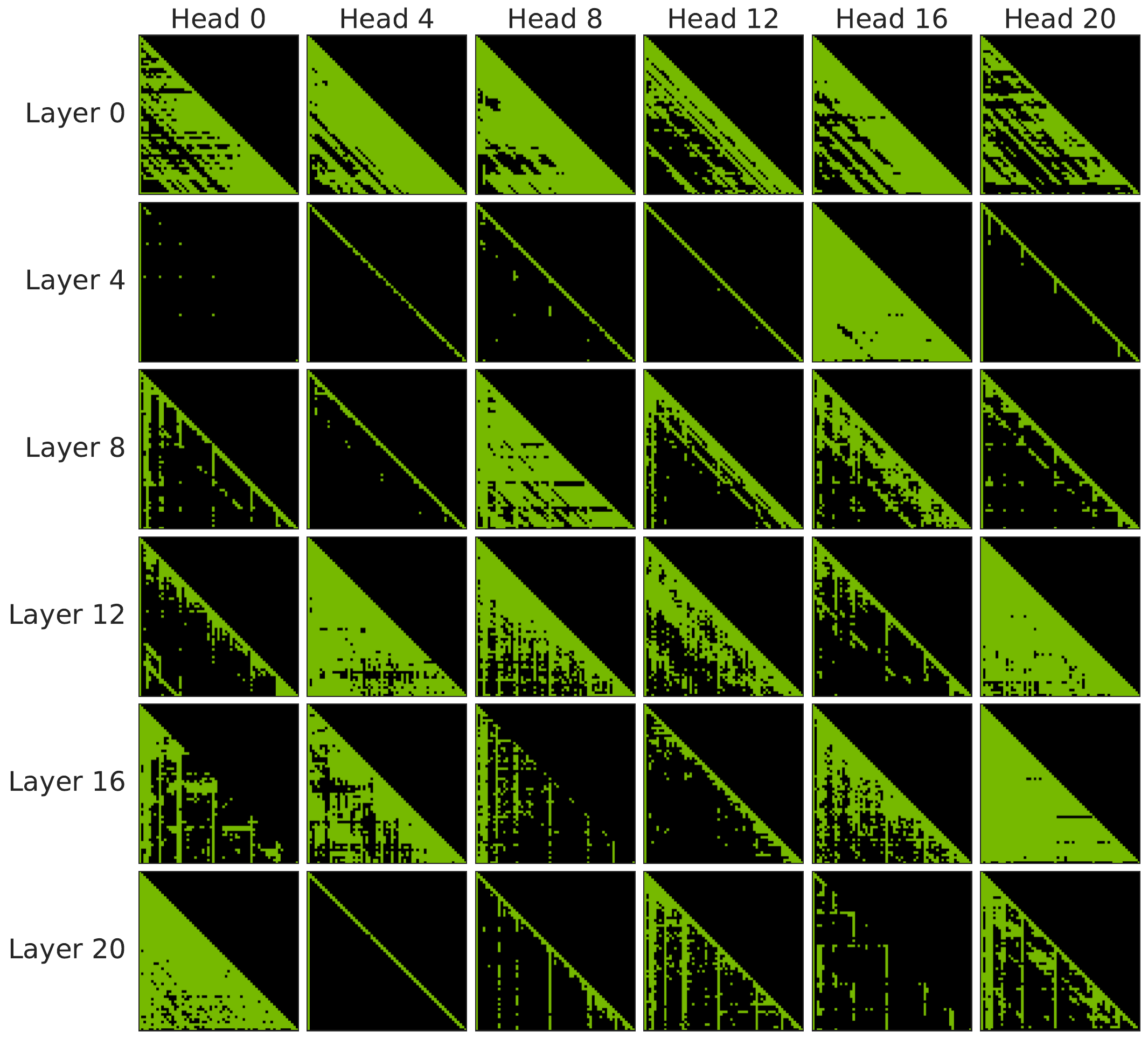}
    \vspace{-2em}
    \caption{Block sparsity distribution across layers and heads for Llama-8B on 8K context. Taken from NIAH benchmark sample with threshold $\lambda=0.03$. Substantial head-level and layer-level variance motivates adaptive thresholding strategies.
    }
    \label{fig:sparsity_dist}
\end{figure}

\textbf{Sparsity Distribution Analysis.} Figure~\ref{fig:sparsity_dist} illustrates how sparsity varies across layers and attention heads, revealing the attention patterns produced by the model. We observe substantial heterogeneity: different layers exhibit different sparsity levels, and individual heads within each layer also show significant variance. Crucially, \algname{} naturally incorporates this heterogeneity without requiring explicit mechanisms like top-k selection or head pruning—by applying the same threshold across all layers and heads, our method automatically adapts to each layer's and head's natural attention distribution, pruning more aggressively where attention is naturally more concentrated and preserving more blocks where attention is more diffuse.

\textbf{Combination with Other Sparsity Methods.} Table~\ref{tab:combination} explores the combination of \algname{} with other attention sparsity techniques. We find that \algname{} can be effectively composed with both prefill-optimized methods (XAttention) and KV cache compression methods (RocketKV). When XAttention (prefill) is combined with \algname{} (decode), accuracy degradation remains minimal, demonstrating that the methods are largely orthogonal. Similarly, combining \algname{} (prefill) with RocketKV maintains strong performance. These results show that \algname{} provides a flexible building block for end-to-end optimization in existing sparse attention pipelines, and show strong potential for composing with other fine-grained channel/head pruning methods~\cite{xu2024think}.

\begin{table}[t]
\centering
\vspace{-.5em}
\caption{Performance of combining \algname{} with other sparsity methods on Qwen 8B. \algname{} can be effectively composed with both prefill-optimized methods (XAttention) and KV cache compression methods (RocketKV), providing flexible deployment options. Numbers in parentheses show change from dense baseline.}
\vspace{.5em}
\label{tab:combination}
\resizebox{\linewidth}{!}{%
\begin{tabular}{llcc}
\toprule
\textbf{Prefill Method} & \textbf{Decode Method} & \textbf{RULER-16K} & \textbf{LongBench-16K} \\
\midrule
Dense Attention & Dense Attention & 93.22 & 29.4 \\
\midrule
XAttention & Dense Attention & 92.99 (-0.23) & 29.1 (-0.3) \\
XAttention & \algname{} & 92.89 (-0.33) & 28.8 (-0.6) \\
\midrule
Dense Attention & RocketKV & 92.72 (-0.50) & 30.0 (+0.6) \\
\algname{} & RocketKV & 92.60 (-0.62) & 29.4 (-0.0) \\
\bottomrule
\end{tabular}%
}
\end{table}

\begin{table}[]
\centering
\caption{Performance on very long sequences with RepoQA benchmark. We evaluate \algname{} on code repository understanding tasks at 16K and 200K context lengths. Sparsity (P) and Sparsity (D) denote \textit{achieved sparsity} in the prefill and decode phases.}
\vspace{.5em}
\label{tab:repoqa}
\resizebox{\columnwidth}{!}{%
\begin{tabular}{lcccc}
\toprule
\textbf{Context} & \textbf{Attention Mode} & \textbf{Sparsity (P)} & \textbf{Sparsity (D)} & \textbf{Accuracy} \\
\midrule
\multicolumn{5}{c}{\textit{Qwen3-Coder-30B-A3B-Instruct, 16K Context}} \\
\midrule
16K & Full (Dense) & 0\% & 0\% & 0.897 \\
16K & \algname{} Prefill & 64.1\% & 0\% & 0.904 \\
16K & \algname{} Prefill+Decode & 64.1\% & 48.4\% & 0.882 \\
\midrule
\multicolumn{5}{c}{\textit{Qwen3-Coder-30B-A3B-Instruct, 200K Context}} \\
\midrule
200K & Full (Dense) & 0\% & 0\% & 0.850 \\
200K & \algname{} Prefill & 57.5\% & 0\% & 0.841 \\
200K & \algname{} Prefill+Decode & 57.5\% & 40.8\% & 0.838 \\
\bottomrule
\end{tabular}%
}
\end{table}

\textbf{Very Long Sequence Lengths.} We evaluate \algname{} on extremely long sequences using the RepoQA benchmark~\cite{liu2024repoqa}. Table~\ref{tab:repoqa} presents results on Qwen3-Coder-30B at 16K and 200K context lengths. At 200K tokens, \algname{} achieves a high prefill sparsity ($\sim$58\%) with a minimal accuracy drop, and applying sparsity to both prefill and decode phases provides additional computational savings with negligible incremental cost. Notably, longer contexts exhibit higher natural sparsity, making our method increasingly effective for extreme-length scenarios where dense attention becomes impractical.

\begin{figure}[]
    \centering
    \includegraphics[width=0.9\columnwidth]{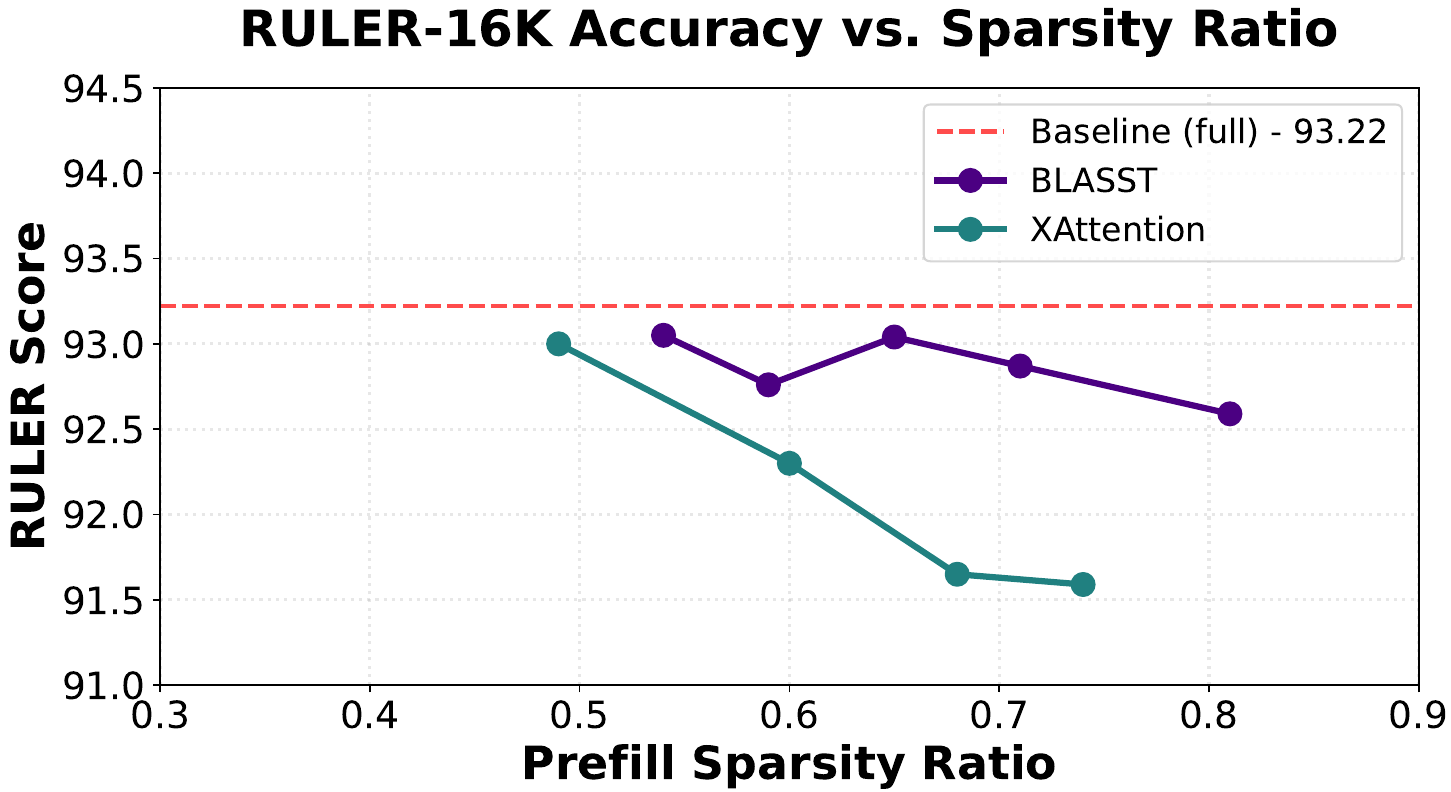}
    \vspace{-1em}
    \caption{Accuracy-sparsity trade-off at high achieved sparsity levels on RULER-16K for Qwen3-8B. \algname{} shows more stable degradation compared to XAttention, maintaining better accuracy at aggressive sparsity settings. This shows the effectiveness of using actual softmax statistics versus proxy-based importance scores.}
    \label{fig:extreme_sparsity}
    \vspace{-1em}
\end{figure}

\textbf{Extreme Sparsity Analysis.} Figure~\ref{fig:extreme_sparsity} shows \algname{}'s behavior at higher sparsity levels (70--90\%) on RULER benchmarks. Compared to XAttention, \algname{} shows more stable accuracy degradation across increasing sparsity levels. Although XAttention shows sharper accuracy drops at high sparsity, \algname{}'s threshold-based pruning using actual softmax statistics (rather than proxy scores) enables more graceful degradation. This stability makes \algname{} more suitable for aggressive sparsity settings where computational efficiency is critical.

\textbf{Tile Row Reordering.} We also investigate whether permuting the tile-row processing order could improve pruning accuracy by establishing a better running maximum earlier. Results in Appendix~\ref{sec:appendix_additional} (Figure~\ref{fig:reorder}) show that the effect is dataset-dependent but generally negligible, confirming \algname{}'s robustness to processing order.

\section{Conclusion}

We presented \algname{}, a simple yet effective sparse attention method that dynamically prunes attention computations by reusing online softmax statistics. \algname{} is easy to adopt: it requires no pre-computation or training, supports both prefill and decode phases, is optimized for modern hardware, and is already integrated into multiple inference frameworks. By substantially accelerating the attention mechanism with minimal accuracy degradation, \algname{} makes long-context inference significantly more practical. Our automated calibration and sparsity-aware training further enhance its robustness and flexibility, providing a practical foundation for efficient long-context transformers.

Looking forward, we believe that the combination of hardware-aware sparse patterns, learned sparsity through training, and adaptive hybrid methods will be the key to unlocking the full potential of future agentic AI systems.

\section*{Acknowledgments}

The authors thank InnoMatrix for providing cloud compute resources for kernel benchmarking.

\bibliography{example_paper}
\bibliographystyle{mlsys2025}



\clearpage
\appendix

\section{Additional Experimental Results}
\label{sec:appendix_additional}

\subsection{Large Model Evaluations}

To evaluate the scalability and robustness of our method, we measured performance across long-context summarization and retrieval tasks. As shown in Table~\ref{tab:large_model_longbench} and Table~\ref{tab:large_model_niah}, our method maintains baseline accuracy at extreme sparsity (70--80\%) using larger models like Qwen3-30B-A3B-Instruct and Llama-3.1-70B-Instruct.

\begin{table}[H]
    \centering
    \caption{Impact of sparsity on LongBench performance using Qwen3-30B-A3B-Instruct. BLASST maintains accuracy comparable to the dense baseline (0.0 sparsity) even as sparsity increases to 70\%, demonstrating robustness in long-context summarization.}
    \vspace{.5em}
    \label{tab:large_model_longbench}
    \footnotesize
    \begin{tabular}{ccc}
        \toprule
        \multirow{2}{*}{\textbf{Target Sparsity}} & \textbf{LongBench V1} & \textbf{LongBench V2} \\
        \textbf{} & \textbf{Overall Accuracy} & \textbf{Overall Accuracy} \\
        \midrule
        0\% & 47.77 & 36.28 \\
        50\% & 47.43 & 38.14 \\
        60\% & 47.47 & 39.53 \\
        70\% & 47.21 & 39.53 \\
        80\% & 46.50 & 37.21 \\
        90\% & 45.97 & 37.21 \\
        \bottomrule
    \end{tabular}
\end{table}

\begin{table}[H]
    \centering
    \caption{Accuracy on the RULER hard subset using Llama-3.1-70B-Instruct. The method retains $>97\%$ accuracy on needle-in-a-haystack tasks even at aggressive sparsity levels (up to 80\%), confirming effective information retention.}
    \vspace{.5em}
    \label{tab:large_model_niah}
    \footnotesize
    \begin{tabular}{ccc}
        \toprule
        \textbf{Target Sparsity} & \textbf{RULER-hard-8k} & \textbf{RULER-hard-16k} \\
        \midrule
        0\% & 97.40\% & 99.06\% \\
        20\% & 97.38\% & 98.98\% \\
        40\% & 97.31\% & 98.80\% \\
        60\% & 97.20\% & 98.59\% \\
        80\% & 97.07\% & 98.28\% \\
        \bottomrule
    \end{tabular}
\end{table}

\subsection{MLA Compatibility}

Table~\ref{tab:deepseek_accuracy} demonstrates that \algname{} is highly compatible with Multi-Head Latent Attention (MLA). When evaluating DeepSeek-R1 NVFP4 on GPQA Diamond, MMLU Pro, and LiveCodeBench, the model maintains near-baseline accuracy even at 60\% sparsity.

\begin{table}[H]
\centering
\setlength{\tabcolsep}{4pt}
\caption{DeepSeek-R1 NVFP4 using BLASST evaluated on GPQA Diamond, MMLU Pro, and LiveCodeBench at different \textit{target sparsity} levels. Minimal accuracy degradation demonstrates that \algname{} is compatible with MLA.}
\vspace{.5em}
\footnotesize
\label{tab:deepseek_accuracy}
\begin{tabular}{lccccc}
\toprule
\textbf{Sparsity} & \textbf{GPQA Diamond} & \textbf{MMLU Pro} & \textbf{LiveCodeBench}\\
\midrule
0\%  & 0.7071 & 0.8302 & 0.5735 \\
50\% & 0.7121 & 0.8283 & 0.5691 \\
60\% & 0.7109 & 0.8266 & 0.5677 \\
\bottomrule
\end{tabular}
\end{table}

\subsection{Calibration Stability Across Datasets}

We evaluate whether the calibrated parameter $a$ transfers across different task types. Table~\ref{tab:dataset_metrics} reports the achieved sparsity when calibrating on individual dataset subsets with a target sparsity of 50\%; this per-dataset breakdown is for illustration purposes only, as in practice we calibrate on a combined, diverse sample dataset. For prefill, all datasets maintain similar achieved sparsity levels, confirming cross-task stability. For decode, two datasets (niah\_single and qa) yield noticeably lower $a$ values; both tasks involve retrieval-focused decoding where the model attends narrowly to specific relevant spans, producing inherently more concentrated attention distributions that require a smaller threshold to reach the target sparsity. Despite this task-dependent variation in $a$, the achieved sparsity remains close to 50\% across all datasets, confirming that a single calibration on a mixed dataset is sufficient for robust deployment across diverse workloads.

\begin{table}[H]
    \centering
    \caption{Calibration stability across diverse datasets on Llama-3.1-8B with a target sparsity of 50\%. We report the calibrated parameter $a$ (where $\lambda=a/L$) and the resulting \textit{achieved sparsity} for both prefill and decode phases. Similar parameter values across tasks confirms that \algname{} generalizes without task-specific tuning.}
    \vspace{.5em}
    \label{tab:dataset_metrics}
    \begin{tabular}{lcccc}
        \toprule
        \multirow{2}{*}{\textbf{Dataset}} & \multicolumn{2}{c}{\textbf{Prefill}} & \multicolumn{2}{c}{\textbf{Decode}} \\
        \cmidrule(lr){2-3} \cmidrule(lr){4-5}
         & $\mathbf{a}$ & \textbf{sparsity} & $\mathbf{a}$ & \textbf{sparsity} \\
        \midrule
        niah\_single      & 920  & 49.98\% & \textit{4.6}  & 49.11\% \\
        niah\_multikey    & 1099 & 46.89\% & 11.4 & 47.87\% \\
        niah\_multivalue  & 1012 & 47.82\% & 11.7 & 48.17\% \\
        niah\_multiquery  & 1100 & 46.38\% & 9.3  & 48.15\% \\
        cwe               & 1020 & 46.69\% & 10.8 & 51.04\% \\
        qa                & 900  & 48.68\% & \textit{5.4}  & 50.97\% \\
        \bottomrule
    \end{tabular}
\end{table}

\subsection{Tile Row Reordering}

We investigated whether permuting the tile-row processing order could improve pruning accuracy. This is motivated by the phenomenon observed in StreamingLLM~\cite{xiao2023efficient}, where recent tokens at the end (local window) and sink tokens at the beginning of the sequence tend to have high attention scores. By processing tiles containing the local window first, the running maximum $m_i$ can be quickly populated with these high-scoring tokens, establishing a better proxy for the global maximum earlier in the computation. This enables more accurate skip decisions for subsequent blocks. Importantly, \algname{} supports such reordering flexibility at the kernel scheduling level with negligible overhead.

\begin{figure}[H]
 \centering
 \includegraphics[width=\columnwidth]{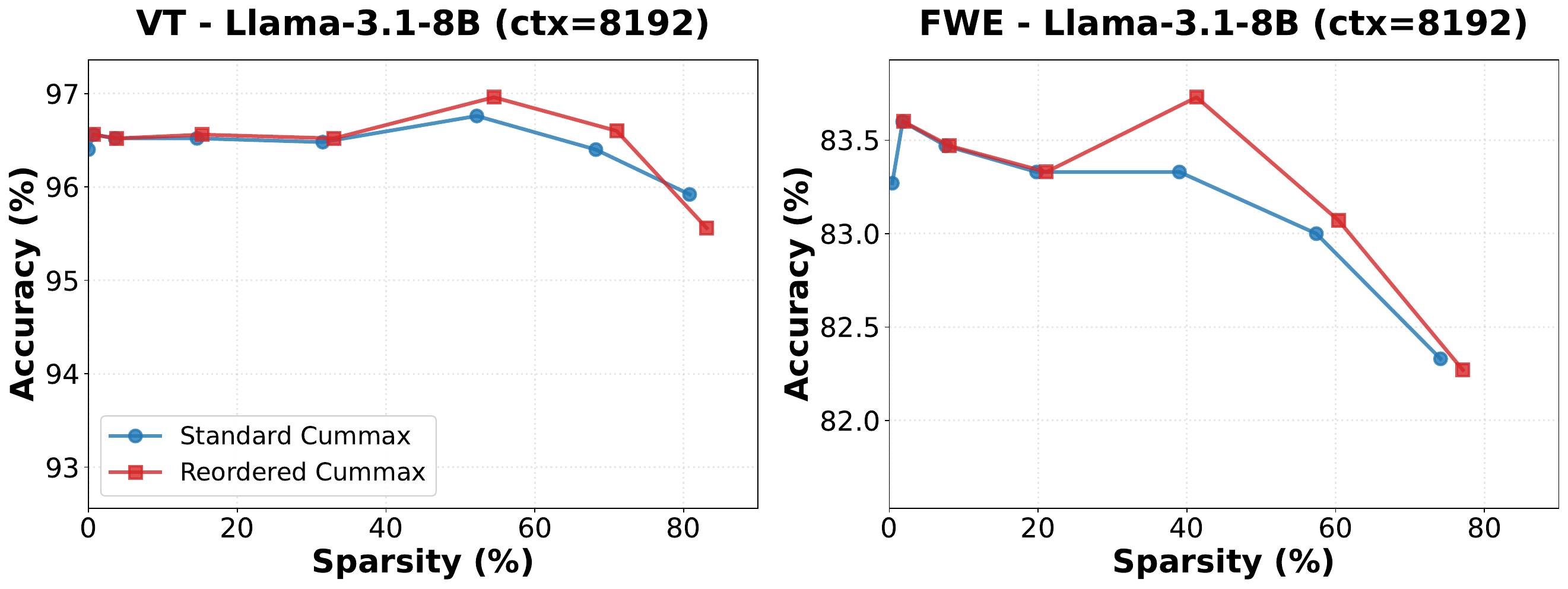}
 \vspace{-1em}
 \caption{Effect of tile row reordering on the accuracy-sparsity trade-off for Llama 3.1 8B (ctx=8192). We compare \emph{Standard Cummax} (processing tiles sequentially) with \emph{Reordered Cummax} (processing tiles in reverse order). The plots for both VT and FWE benchmarks show that reordering has a negligible impact on model accuracy at a given sparsity level.}
 \label{fig:reorder}
\end{figure}

Figure~\ref{fig:reorder} compares standard sequential processing against reordered processing on VT and FWE tasks. The results show dataset-dependent behavior: reordering yields similar performance on VT but provides noticeable improvements on FWE\@. This suggests that the effectiveness of reordering largely depends on the specific attention patterns of each dataset. Nevertheless, this demonstrates a valuable property of \algname{}: the algorithm is robust to different processing orders and can accommodate various optimization strategies. The flexibility to support tile reordering shows the potential for dataset-specific optimizations without requiring fundamental algorithmic changes.

\section{Error Bound Analysis}

We derive an error bound for the output approximation introduced by skipping attention blocks in \algname{}.

Consider a single query token with attention output
\[
y = \frac{\sum_{j=1}^{T_c} \sum_{k=1}^{B_c} \exp(s_{jk} - M)\, v_{jk}}{Z},
\]
where $B_c$ is the KV block size, $s_{jk}$ are the attention scores, $M = \max_{j,k} s_{jk}$ is the global maximum, $Z = \sum_{j,k} \exp(s_{jk} - M)$ is the softmax normalization constant, and $v_{jk}$ are the value vectors.

\textbf{Per-block mass bound.}
When \algname{} skips block $j$, the criterion $\tilde{m}^{(j)} - m^{(j)} < \ln \lambda$ guarantees $\exp(\tilde{m}^{(j)} - m^{(j)}) < \lambda$, where $\tilde{m}^{(j)}$ is the block-local maximum score and $m^{(j)}$ is the running maximum.
Since $m^{(j)} \leq M$, every score in a skipped block satisfies
\[
\exp(s_{jk} - M) \;\leq\; \exp\!\bigl(\tilde{m}^{(j)} - M\bigr) \;\leq\; \exp\!\bigl(\tilde{m}^{(j)} - m^{(j)}\bigr) \;<\; \lambda.
\]
Summing over all $B_c$ tokens in the block, the total unnormalized attention mass of a single skipped block is
\[
\sum_{k=1}^{B_c} \exp(s_{jk} - M) \;<\; B_c \cdot \lambda.
\]

\textbf{Output error bound.}
Let $\mathcal{S}$ denote the set of skipped blocks and let $V_{\max} = \max_{j,k} \|v_{jk}\|$.
Since $Z \geq 1$ (the element achieving the global maximum contributes $\exp(0) = 1$), each skipped token's softmax weight satisfies
\[
p_{jk} \;=\; \frac{\exp(s_{jk} - M)}{Z} \;<\; \lambda.
\]
The output error equals the total contribution of skipped tokens:
\[
\|y - \hat{y}\| \;=\; \left\|\sum_{j \in \mathcal{S}} \sum_{k=1}^{B_c} p_{jk}\, v_{jk}\right\| \;\leq\; \underbrace{\left(\sum_{j \in \mathcal{S}} \sum_{k=1}^{B_c} p_{jk}\right)}_{\delta} V_{\max}.
\]
Each skipped token contributes at most $\lambda\, V_{\max}$ to this sum. Aggregating over all $|\mathcal{S}|$ skipped blocks,
\[
\boxed{\|y - \hat{y}\| \;\leq\; \delta\, V_{\max} \;<\; |\mathcal{S}|\, B_c\, \lambda\, V_{\max}.}
\]
In practice, because the approximate output $\hat{y}$ is renormalized over non-skipped blocks only (denominator $Z - Z_{\mathcal{S}}$ instead of $Z$), a correction of order $\delta^2 V_{\max}$ arises; this is negligible since $\delta \ll 1$.

\section{Artifact Appendix}
\subsection{Abstract}

This artifact evaluation provides the framework and code necessary to reproduce the kernel-level performance benchmarks for \algname{}. The repository focuses on evaluating our custom kernels against a SOTA baseline across both prefill and decode phases. Utilizing automated sweeps across various threshold scale factors, the provided scripts systematically measure exact attention sparsity percentages, execution times, memory bandwidth, and speedups compared to dense baselines. Our work has been integrated into TensorRT-LLM and FlashInfer, and we pull the relevant code from these sources for evaluation. The framework is designed to target and benchmark performance on NVIDIA Hopper (H200) and Blackwell (B200) architectures within a containerized Docker or Singularity environment, handling all necessary installation and measurement.

\subsection{Artifact check-list (meta-information)}


{\small
\begin{itemize}
  \item {\bf Algorithm:} \algname{} (Skip-Softmax)
  \item {\bf Compilation:} CUDA nvcc builds for kernel templates
  \item {\bf Binary:} Some closed binaries used to measure sparsity
  \item {\bf Run-time environment:} Docker
  \item {\bf Hardware:} H200 and B200 GPUs, many-core x86 CPU, SSD
  \item {\bf Execution:} Python and bash scripts
  \item {\bf Metrics:} Skipping threshold, sparsity, throughput, memory bandwidth
  \item {\bf Output:} Standard output (stdout)
  \item {\bf Experiments:} Single GPU kernel benchmarks and sparsity data collection.
  \item {\bf How much disk space required:} 100~GB
  \item {\bf How much time is needed to prepare workflow:} 45 minutes
  \item {\bf How much time is needed to complete experiments:} 1 hour
  \item {\bf Publicly available:} Yes
  \item {\bf Code licenses:} Apache 2.0
  \item {\bf Workflow framework used:} TensorRT-LLM, FlashInfer
  \item {\bf Archived:} TBD
\end{itemize}
}

\subsection{Description}






\subsubsection{How delivered}
The artifact is delivered as an open-source GitHub repository. It can be obtained by cloning the repository and its external submodules via \texttt{git clone} \url{git@github.com:cameronshinn/blasst-ae-mlsys26.git} \texttt{--recursive}.

\subsubsection{Hardware dependencies}
The evaluation requires a host machine equipped with a many-core x86 CPU, an NVIDIA Hopper (H200) GPU, or an NVIDIA Blackwell (B200) GPU (depending on which kernels you want to evaluate). The host system should also have an SSD with approximately 100~GB of available storage space to accommodate the required container images, compiled binaries, and generated benchmark data.

\subsubsection{Software dependencies}
The artifact relies on a containerized run-time environment. The host system must have either Docker (with the NVIDIA Container Toolkit installed) or Singularity available. The provided startup scripts automatically pull and utilize the official TensorRT-LLM release container (\url{nvcr.io/nvidia/tensorrt-llm/release:1.3.0rc6}). A compatible Linux host distribution with up-to-date NVIDIA drivers supporting the target Hopper or Blackwell architectures is required.

\subsubsection{Data sets}
The core kernel benchmarks sweep across various threshold scale factors, evaluating throughput, memory bandwidth, and execution time on randomly initialized tensors. The artifact also includes closed sm100 binaries used to measure and collect exact sparsity percentages dynamically during execution.

\subsection{Installation}

To install and prepare the artifact, first clone the repository along with its required submodules:

\begin{lstlisting}
git clone git@github.com:cameronshinn/blasst-ae-mlsys26.git --recursive
\end{lstlisting}

Next, initialize the containerized environment. The repository provides a convenience script to automatically launch the required Docker container (falling back to Singularity if Docker is unavailable) and mount the repository to the \texttt{/workspace} directory:

\begin{lstlisting}
./start_docker.sh
cd /workspace
\end{lstlisting}

\subsection{Experiment workflow}

The evaluation workflow is organized by target hardware architecture. For the NVIDIA Hopper architecture, the workflow is further decoupled by attention phase into separate directories. For the NVIDIA Blackwell architecture, both prefill and decode evaluations are consolidated into a single directory. After launching the container and navigating to \texttt{/workspace}, the general workflow proceeds as follows:

\begin{enumerate}
  \item Navigate to the specific subdirectory corresponding to the available architecture and desired evaluation phase (e.g., \path{hopper_prefill}, \path{hopper_decode}, or \path{blackwell}).
  \item Follow the steps in the \texttt{README.md} file of that specific subdirectory to compile the kernels and initiate the automated benchmarks.
  \item The script will automatically sweep across various threshold scale factors, executing both the \algname{} kernels and the dense SOTA baselines.
  \item Collected measurements for sparsity, execution time, and memory bandwidth, will be logged directly to standard output.
\end{enumerate}

\subsection{Evaluation and expected result}

We expect our results to align with what's shown in Table~\ref{tab:combined_speedup}.
Each \texttt{README.md} file in the folders of our repository contain expected outputs of their associated scripts.


\end{document}